%% file: main.tex
\documentclass[10pt,twocolumn,letterpaper]{article}
\usepackage[pagenumbers]{paperstyle}

\input{macros}

\definecolor{linkblue}{rgb}{0.21,0.49,0.74}
\usepackage[pagebackref,breaklinks,colorlinks,allcolors=linkblue]{hyperref}
\hypersetup{pdftitle={How Many Samples Is RL Worth? Best-of-N versus RL Post-Training for Text-to-Image Models},
  pdfauthor={Salil Bhatnagar}}

\newif\ifflux
\fluxtrue
\input{flux_numbers}

\title{How Many Samples Is RL Worth?\\ Best-of-$N$ versus RL Post-Training for Text-to-Image Models}
\author{Salil Bhatnagar\\
Bosch Research India\\
{\tt\small salil.bhatnagar@in.bosch.com}}

\begin{document}
\crefname{appendix}{Supp.~Sec.}{Supp.~Secs.}\Crefname{appendix}{Supplementary Section}{Supplementary Sections}
\maketitle
\input{sections/00_abstract}
\input{sections/01_intro}

\input{sections/02_related}
\input{sections/03_setup}
\input{sections/04_method}
\input{sections/05_results}
\input{sections/06_analysis}
\input{sections/07_limitations}
\input{sections/08_conclusion}
\input{sections/09_llm_statement}

{
    \small
    \bibliography{refs}
}

\input{sections/A_supplement}
\end{document}

%% file: macros.tex
\usepackage{amsthm,bm,multirow,microtype,array}

\newcolumntype{L}[1]{>{\raggedright\arraybackslash}p{#1}}

\newcommand{\base}{\textsc{Base}\xspace}                             %
\newcommand{\bstar}{\textsc{Base}\ensuremath{^\star}\xspace}         %
\newcommand{\bzero}{\textsc{Base}\ensuremath{_{1}}\xspace}           %
\newcommand{\grpo}{\textsc{GRPO}\xspace}                             %
\newcommand{\nft}{\textsc{NFT}\xspace}                               %
\newcommand{\txt}{\textsc{GRPO-Text}\xspace}                         %

\newcommand{\vlm}{\mathrm{VLM}}

\newcommand{\passk}[1]{\mathrm{pass@}#1}
\newcommand{\bon}[1]{\mathrm{BoN@}#1}
\newcommand{\nstar}{N^\star}
\newcommand{\KL}{\mathrm{KL}}
\newcommand{\logit}{\operatorname{logit}}

\newcommand{\ci}[2]{{\scriptsize(#1,\,#2)}}
\newcommand{\m}{\ensuremath{-}}
\newcommand{\p}{\ensuremath{+}}

\newtheorem{proposition}{Proposition}

%% file: flux_numbers.tex
\newcommand{\fluxGPU}{NVIDIA~L40}
\newcommand{\fluxPrompts}{398}
\newcommand{\fluxPromptText}{all 398 test prompts}
\newcommand{\fluxSeedText}{32 seeds for the base model and 16 for the adapter}
\newcommand{\fluxBaseOne}{46.0}
\newcommand{\fluxRLOne}{70.3}
\newcommand{\fluxNstarPerfect}{6.4 (5.1 to 8.1)}
\newcommand{\fluxBonSel}{62.1}
\newcommand{\fluxD}{$-$8.1 ($-$10.9 to $-$5.4)}
\newcommand{\fluxDdet}{$-$9.2}
\newcommand{\fluxDvlm}{$+$0.3}

%% file: sections/00_abstract.tex
\begin{abstract}
Reinforcement learning (RL) on a reward that can be checked automatically has become a common last step in the
training of text-to-image models. A simpler way to get more correct images is best-of-$N$: draw $N$ images from the
base model and keep the one that a picker prefers. We ask how many base model samples one RL sample is worth, and
call this number the exchange rate. We measure it for Stable Diffusion 3.5 Medium and its public Flow-GRPO checkpoint
on 398 new prompts in the style of GenEval, the benchmark behind the RL reward, with human labels on every prompt and
two automatic judges other than the reward. The answer depends on the prompt. Four fifths of the prompts do not ask
for a position (for example, one object left of another). On these, the base model catches up: best-of-16 with an
off-the-shelf vision-language model (VLM) as the picker is about as good as one RL image. A person who looks through
up to 3 base images finds a correct one as often as one RL image is correct. On the prompts that ask
for a position, RL is far ahead: a person finds a correct image among 8 base images for only 52\% of the prompts,
against 92\% for one RL image. Over all prompts, our primary test, which we
fixed in advance, finds best-of-16 with the VLM picker 9.3 points behind one RL image when both automatic judges must
pass an image. People see a smaller gap, 5.5 and 6.5 points in two checks. FLUX.1-dev shows a similar
gap in the same test. RL also solves 20 prompts that the base model misses in 128 tries, and 1,024 tries recover
half of them. The automatic judges are wrong
in opposite directions: a detector-based judge like the reward, as used in our primary test, reports a larger gap
than people see, and VLM judges report little or none. Comparisons of RL with best-of-$N$ should therefore check the
selected images by hand.
\end{abstract}

%% file: sections/01_intro.tex
\section{Introduction}
\label{sec:intro}

\begin{figure*}[t]
\centering
\includegraphics[width=\textwidth]{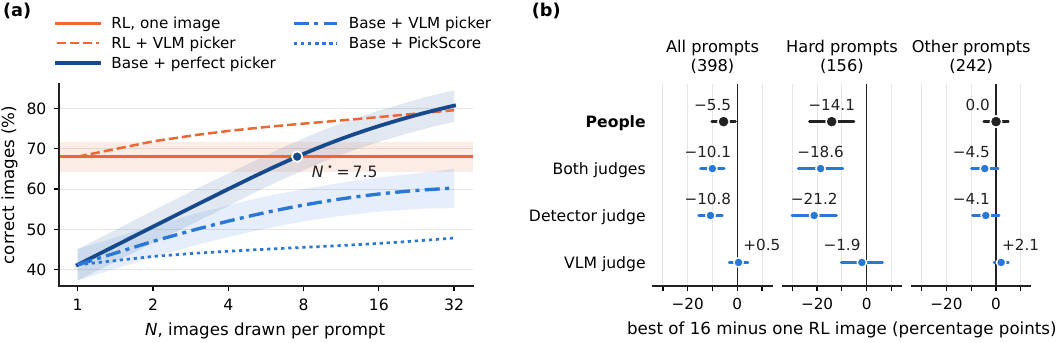}
\caption{\textbf{One RL image against best-of-$N$ of the base model, on 398 test prompts.}
(a)~Share of correct images when the base model draws $N$ images per prompt and a picker keeps one. With a perfect
picker, which finds a correct image whenever there is one, the base model matches one RL image (flat orange line) at
$\nstar=7.5$. With pickers that can be run in practice it stays below the line up to $N=32$; the VLM picker also
helps the RL model (dashed). The dotted curve is best-of-$N$ with PickScore as the picker. An image is correct when both automatic judges pass it; bands are 95\% intervals.
(b)~Best-of-16 with the VLM picker minus one RL image, on 796 images that people judged in our first check, two
per prompt; below 0, RL is better. People see a gap of 5.5 points, the detector judge about twice as much, the VLM
judge none. Hard: either model passes at most 4 of its 32 images.}
\label{fig:teaser}
\end{figure*}

Reinforcement learning (RL) on a reward that can be checked automatically has become a common last step in the
training of text-to-image models~\cite{black2024ddpo,fan2023dpok,liu2025flowgrpo,xue2025dancegrpo,zheng2026diffusionnft},
and the gains on compositional benchmarks are large. Flow-GRPO~\cite{liu2025flowgrpo}, for example, fine-tunes Stable
Diffusion 3.5 Medium (SD3.5-M)~\cite{esser2024sd3} with the GenEval detector~\cite{ghosh2023geneval} as its reward and
raises the GenEval score from 0.63 to 0.95.

There is, however, a much simpler way to get more correct images from the base model. One can draw several images
for the same prompt and keep the one that a picker likes best, a method called best-of-$N$. With a
vision-language model (VLM) as the picker, best-of-2048 raises SANA~1.5 from 0.81 to 0.96 on
GenEval~\cite{xie2025sana15}, which is about where RL brings SD3.5-M. A practitioner therefore has a real choice. RL is
paid for once, during training, and adds nothing to the cost of an image. Best-of-$N$ needs no training, but every
image costs about $N$ times more. In this paper we put a number on this choice. We ask how many base model samples one
RL sample is worth, and we call this number the \emph{exchange rate} $\nstar$: the smallest $N$ at which best-of-$N$ of
the base model is as accurate as one RL image.

The exchange rate also says something about what RL does to a model. For language models, RL with verifiable rewards
wins when a single answer is drawn, but the base model catches up when both models are given many
attempts~\cite{yue2025rl}. This suggests that RL mostly makes the base model's own answers more likely. If the same
holds for images, one RL sample is worth $\nstar$ base samples and no more. If RL instead reaches prompts that the base
model cannot solve, no value of $N$ is enough for those prompts. Concurrent work~\cite{wang2026quantile} finds that the
base model catches up with RL checkpoints of SD3.5-M at large $N$ when the evaluation metric itself picks the best
image. In practice the metric is not available as a picker, and the answer then changes.

We study SD3.5-M and its public Flow-GRPO checkpoint, and we add FLUX.1-dev,
DiffusionNFT~\cite{zheng2026diffusionnft} and a text-rendering reward as further cases. We use 398 test prompts that
the RL model never saw in training. Of these, 258 ask for harder compositions of the objects that its reward knows,
and 140 use everyday objects that it does not know. All of them follow the prompt templates of GenEval, on which the
RL model was trained.

Three choices keep the comparison fair. First, in our main comparisons a picker that can be deployed, which we
call a practical picker, is never the same model as a judge. Only the perfect picker uses the judges to pick, and it
cannot be deployed. Second, we do not trust a single judge. We use two automatic judges of different kinds, a VLM and an
open-vocabulary detector, and people judged images for all 398 prompts in three checks: two of best-of-16
against one RL image and one of the perfect picker.
Third, the main choices were fixed in a written plan before any test image was generated: the guidance scale
of the base model, the picker, the budget of 16 samples and the single primary test. We call this plan the registered plan,
and its test the registered test. The plan is a time-stamped commit in a private repository, not an entry in a public
registry. The human checks are not part of it (\cref{sec:setup}).

Our first finding is that the answer depends on the prompt. Four fifths of our prompts do not ask for a
position. On these, a few base samples are enough. When a person looks for a correct image, one of the first 3
base images is correct about as often as one RL image is. A perfect picker needs about 4 base samples by our
judges, and best-of-16 with an off-the-shelf VLM picker is about as good as one RL image. There, much of what RL adds
is already within reach of the base model.

The other fifth of the prompts ask for a position, for example that one
object is left of another. On these, RL is far ahead. A person finds a correct image among 8 base images for only
52\% of the prompts, where one RL image is correct for 92\% (\cref{fig:person}), and by our judges even 32 base
samples are not enough. Both results in which a person looks for the image come from one part of our third human
check, in which one person per prompt looked through the base images one after another. This may favour the base
model somewhat (\cref{sec:results-rate,sec:supp-search}). But all three human
checks, with four ways of picking a base image, show the same two cases (\cref{tab:together}). Averaged over all prompts,
the two cases give an exchange rate of about 8 with a perfect picker (\cref{fig:teaser}a), in line with the
concurrent result.

Our second finding concerns pickers that can be run in practice, and who judges their choice
(\cref{fig:teaser}b). Best-of-16 with an off-the-shelf VLM picker of 8B parameters, which asks one yes/no question
per image, is 9.3 points behind one RL image over all prompts in our registered test, which counts an image as
correct only when both automatic judges pass it. FLUX.1-dev shows a similar gap by the same two judges; no person
checked its images. For SD3.5-M, people confirm the gap but find it smaller: 5.5 and 6.5 points in two checks. The automatic judges are
wrong in opposite directions. The detector judge, which our registered test also uses, is of
the same kind as the reward that RL was trained on. It misses more of the base model's correct images than of the RL
model's and reports a larger gap than people see, almost twice as large in the first check. The VLM judge is of the same kind as the picker. It accepts
more of the picker's mistakes and reports no gap at all. A comparison of RL with best-of-$N$ that rests on one
judge can go either way.

People also show where this gap is. Almost all of it sits on the 156 prompts that are hard for one model or both,
and most of it on the prompts that ask for a position. With a picker almost five times larger, the gap that people see in the second
check narrows to 2.3 points, but part of this gain may be chance (\cref{sec:limitations}).

Finally, we ask how far RL reaches and what each option costs. In 128 tries per prompt, the RL model solves 20 prompts
that the base model does not solve, and the base model solves 3 that the RL model does not. Given 1,024 tries on these
20 prompts, the base model solves half of them. So RL mostly turns rare successes into frequent ones, but for some
prompts sampling is not a realistic substitute. Best-of-16 costs about 16 times as much per prompt as one RL
image, so 1,000 GPU-hours of RL training are paid back after about 134,000 served prompts.

%% file: sections/02_related.tex
\section{Related work}
\label{sec:related}

\paragraph{RL for diffusion and flow models.}
DDPO and DPOK fine-tune diffusion models on a reward with policy gradients over the denoising
steps~\cite{black2024ddpo,fan2023dpok}, and other methods backpropagate the reward through the
sampler~\cite{clark2024draft,prabhudesai2023alignprop,xu2023imagereward}. Flow-GRPO~\cite{liu2025flowgrpo} and
DanceGRPO~\cite{xue2025dancegrpo} make the flow matching sampler stochastic so that group relative policy optimisation~\cite{shao2024deepseekmath}
can be applied, and many variants and design studies followed~\cite{li2025mixgrpo,he2025tempflowgrpo,choi2026rethinking}. DiffusionNFT and related methods
instead learn on the forward process~\cite{zheng2026diffusionnft,xue2025awm}. Several studies report that a proxy
reward can rise while image quality falls~\cite{wang2026grpoguard,he2025gardo,hong2026hacking,zhang2024tdpo} and that RL reduces the
diversity of what a model draws~\cite{liu2026drift,liu2025diversegrpo}. Most of these works evaluate one image per
prompt. We measure the value of public checkpoints in units of base model samples.

\paragraph{Best-of-$N$ and inference-time scaling.}
Selecting the best of several candidates makes diffusion models more faithful to the prompt~\cite{karthik2023select},
searching over the starting noise helps more than adding sampling steps~\cite{ma2025inference}, and best-of-$N$ with a
VLM verifier raises SANA~1.5 to 0.96 on GenEval~\cite{xie2025sana15}. Particle and tree-based
samplers~\cite{singhal2025fk,kim2025das,jain2025dts} and methods that optimise the starting
noise~\cite{eyring2024reno,zhou2025golden} spend more compute at inference to raise the reward, and some of them match
fine-tuning~\cite{singhal2025fk,kim2025das}. At equal wall-clock time plain
best-of-$N$ matches several of these methods~\cite{rawal2026flashbon}, but it can overfit to its
picker~\cite{huang2025bestofn,khalaf2025hacking}, a weak VLM picker may barely beat a random choice~\cite{seo2026judge},
and with an imperfect verifier, resampling a weaker model may never reach a stronger one~\cite{stroebl2024flaws}.
A VLM that answers separate questions about objects, counts and positions is a stronger verifier for such
searches~\cite{fan2026keystone}, and refining an image step by step with a VLM critic beats parallel sampling on
spatial prompts~\cite{jaiswal2026iterative}.

\paragraph{RL against sampling.}
Training compute can be traded against test-time compute~\cite{jones2021scaling,villalobos2023trading,eyring2025hypernoise,zhang2025parm},
also for language models~\cite{cobbe2021verifiers,brown2024monkeys,snell2024scaling}, where best-of-$N$ and RL have been compared on a
learned reward~\cite{nakano2021webgpt,gao2023scaling}. There, RL raises pass@1 but often not pass@$k$ at
large $k$~\cite{yue2025rl,sun2026budgeted}, although longer training can solve new problems~\cite{liu2025prorl}; recent work describes
RL as a reweighting of the base model~\cite{liu2026nofreelunch,huang2025sharpening,karan2026sampling} and finds that it rarely solves problems the base
model misses in all tries~\cite{yang2026reach,wu2025leash}. In theory, best-of-$N$ and KL-regularised RL both reweight the base
distribution~\cite{beirami2025bon,gui2024bonbon,korbak2022rlkl,uehara2024entropy}. For images, stronger guidance and some distillation methods
improve a single sample but lose at large $k$~\cite{wang2026coverage}. The work closest to ours is concurrent: Wang
and Ying~\cite{wang2026quantile} compare RL checkpoints of SD3.5-M with best-of-$N$ of the base model on held-out
prompts and find that the gains of RL shrink or reverse at large $N$. Their picker is the evaluation metric itself,
which corresponds to our perfect picker, and our result for that picker agrees with theirs. An RL method for
SD3.5-M~\cite{zhu2026crd} plots best-of-$N$ on GenEval as a side result, also with the evaluation reward as the picker,
and concludes that RL matches or beats it with one sample. We add a
picker that differs from the judge, human labels, an exchange rate with its cost, and a count of the prompts
that only one model solves.

\paragraph{Evaluating text-image alignment.}
GenEval~\cite{ghosh2023geneval} checks objects, counts, colours and positions with an object detector, and it has
drifted away from human judgement as models improved~\cite{kamath2025geneval2}. VLM-based
metrics~\cite{hu2023tifa,cho2024dsg,lin2024vqascore} are more reliable for some prompt types than for
others~\cite{wiles2025gecko,kasaei2025evaluators} and tend to answer ``yes''~\cite{ross2024metric}. A metric that is
calibrated on single images does not control the error of the image that a search selects~\cite{yin2026ship}, and the
choice of judge can reverse a comparison~\cite{li2026verifierchoice}. In other settings, a small human study found a
smaller gain than a GenEval-style detector had reported~\cite{agarwal2026ledge}, and a detector and a VQA-based
metric moved in opposite directions on spatial prompts~\cite{gonuguntla2026plan}. For this
reason we check by hand the images that best-of-$N$ picked, and not only random images.

%% file: sections/03_setup.tex
\section{Models, prompts and judges}
\label{sec:setup}

\paragraph{Models.}
\base is SD3.5-M~\cite{esser2024sd3} sampled with a classifier-free guidance (CFG)~\cite{ho2022cfg} of 4.5, which is
how the RL authors sample it. The guidance that is best for best-of-$N$ need not be the one that is best for a single
image. We therefore chose the guidance of the base model from $\{1, 2.5, 4.5, 7\}$ on a separate set of dev prompts,
described below, by its best-of-16
accuracy with the VLM picker. This gave 7, and we call this model \bstar. It is the base model in our main
comparisons; guidance 4.5 gives almost the same results. \grpo is the public
Flow-GRPO checkpoint trained on the GenEval reward~\cite{liu2025flowgrpo}. It is a LoRA adapter on \base, trained with
a small KL penalty that keeps it close to \base. \nft is the public DiffusionNFT
checkpoint~\cite{zheng2026diffusionnft}, trained on five rewards, among them GenEval, with the same GenEval training
prompts as \grpo. It needs no guidance, so an image costs half as much, and we also compare it with \bzero, the base
model at CFG 1. \txt is Flow-GRPO trained on an OCR reward, which we use for a second task.
\ifflux
As a second base model, we also compare FLUX.1-dev~\cite{bfl2024flux} with a
public Flow-GRPO adapter trained on the GenEval reward~\cite{chen2025mapreducelora}, on \fluxPromptText{} with
\fluxSeedText.
\fi
All SD3.5-M models use the sampler of the Flow-GRPO and DiffusionNFT evaluations: $512\times512$ pixels and 40 Euler steps
with a time shift of 3. The LoRA weights are merged into the model, so an image of \grpo costs exactly as much as an
image of \base.

\paragraph{Prompts.}
The \emph{known-object prompts} are harder compositions of the COCO objects~\cite{lin2014coco}, which are the objects
known to the GenEval detector and therefore to the reward of \grpo. They ask for a count of five, for two counted
classes, for three objects, for three colour bindings, or for a colour together with a position. None of these
compositions appears among the training prompts of \grpo, but they use the template wording of GenEval, on which \grpo
was trained. The \emph{new-object prompts} follow the six prompt types of GenEval with everyday objects outside COCO,
which the reward cannot recognise. An object is kept only if both judges find it in at least 80\% of its real LVIS
photos~\cite{gupta2019lvis}; of 70 candidate objects, 12 have too few photos, 6 fail this test and 52 are kept. Both sets are split into dev and test prompts.
The test set has 258 known-object and 140 new-object prompts, and the dev set has 52 and 35. As the two sets also
differ in their prompt types, we do not attribute a difference between them to the objects alone.
For the second task we use 200 \emph{text prompts} that ask for a text of 3 to 53 characters. \txt was trained on
texts of at most 29 characters, counting spaces, so half of these prompts ask for longer texts than it saw.

\paragraph{Judges.}
A \emph{judge} decides whether an image is correct, and a \emph{picker} ranks the images generated for one prompt. In our main comparisons, no practical picker is also a judge. We use two automatic judges of different kinds. The \emph{VLM judge} is
Qwen3-VL-8B~\cite{bai2025qwen3vl}. Following TIFA and DSG~\cite{hu2023tifa,cho2024dsg}, each prompt is split into
yes/no and ``how many'' questions, and the answers are read from the next-token probabilities. The \emph{detector
judge} applies the rules of GenEval, with its CLIP colour classifier, to the boxes found by
OWLv2~\cite{minderer2023owlv2}, an open-vocabulary detector that plays no part in the training of any model here. Its
two thresholds were tuned once on 240 hand-labelled dev images. An image counts as correct only if \emph{both} judges
pass it. This is our registered rule, and all accuracies use it unless we say otherwise. Since the two
judges can make different mistakes, we also report the main results under each judge alone, and under people.

\paragraph{Pickers.}
The \emph{perfect picker} knows the verdict of the judges. Its best-of-$N$ accuracy is pass@$N$, the upper bound for
any picker. The \emph{VLM picker} is our main practical picker. It asks InternVL3.5-8B~\cite{wang2025internvl35} a
single question per image, \emph{Does this figure show ``\{prompt\}''? Please answer yes or no.}, and uses
$P(\text{yes})$ as the score, as in VQAScore~\cite{lin2024vqascore}. It comes from a different model family than the
VLM judge, although its language model is also a Qwen3 model. PickScore~\cite{kirstain2023pickapic} and CLIPScore~\cite{hessel2021clipscore} serve as weaker pickers. On
the known-object prompts we also use the \emph{GenEval reward} itself, computed as in the RL code. On the
text prompts the picker is
PaddleOCR~\cite{li2022ppocrv3} with access to the target text, which is the reward of \txt. The judge is the VLM judge,
which reads the text without seeing the target.

\paragraph{Seeds and hard prompts.}
\label{sec:exp}
A seed gives the same starting noise to every model, so all comparisons are paired by seed. Every test prompt
receives seeds 0--31 for every model. Prompts that are hard for either model receive more: if \grpo or \bstar passes a
prompt at most 4 times out of 32, both models receive seeds 32--127 for it. We call these 156 prompts the \emph{hard
prompts} and the remaining 242 the other prompts. In total the main study generated 258,048 images in about 100 A100
GPU-hours.

\paragraph{People.}
Three people labelled images: a first labeller and two researchers who work on vision-language models. None of
them is an author, and none took part in the design of the study or knew its hypotheses. Together they made 6,371 labels. In the human checks most images have one label, and 100 to 133 images of each
check have two. Each labeller worked
alone, the model behind an image was hidden, and the images were shown in random order, except where we say otherwise. On 100 images of the first human
check, the two researchers agree with each other on 98 (Cohen's $\kappa=0.95$), and the first labeller agrees with one
of them on 93 of another 100 ($\kappa=0.84$). The labellers took part voluntarily and agreed to the use of their labels. We asked them for no personal information, and we did not seek ethics review. \Cref{sec:supp-prompts} gives the reasons, the written rules and
all agreement numbers.

\paragraph{Choices fixed in advance.}
The registered plan fixed the following choices on dev data, before any test image was generated: the guidance
of \bstar, the budget $N=16$, the VLM picker and its question, the rule for hard prompts, the thresholds of the
detector judge and the single primary test (\cref{sec:method}). \Cref{sec:supp-rules} lists every rule with
its outcome and every departure from it. The plan also named three tests of the setup that had to pass first. On the official
GenEval prompts we reproduce the published scores of \base, \grpo and \nft within 0.03, scored as in the RL code. The
judges count accurately up to five objects but not six or seven, so no prompt asks for more. And a hand check of 50
test images that the judges passed found all 50 correct.

\paragraph{What the plan does not cover.}
The registered plan does not cover the three human checks (\cref{sec:method}). The first was decided after the registered result,
and we report it as exploratory. Each later check had its own plan, fixed before its images were drawn or labelled,
but after the results of the earlier checks were known. Most of the second part of the third check was labelled after
the last date that its plan had set for the analysis (\cref{sec:supp-search}). The split by position was found in the first check and then
named in the later plans. The comparison of the judges with people in the human checks was not in the registered plan either. The plan of the first
check, written before the two researchers made their first label, named how often each judge passes the images that
people call correct or wrong (\cref{tab:judges}). The further analysis of the judges was not planned
(\cref{sec:supp-other,sec:supp-stats,sec:supp-judge2}).

%% file: sections/04_method.tex
\section{Measuring what RL is worth}
\label{sec:method}

\paragraph{Best-of-$N$.}
\label{sec:method-rate}
For one prompt and one model we have $n$ images, from seeds $0,\dots,n-1$. Image $i$ has a picker score $s_i$ and a
verdict $e_i\in\{0,1\}$ from the judge. Best-of-$N$ draws $N$ of the $n$ images at random, keeps the one with the
highest score, and is correct if that image is correct. We compute its expected accuracy $\bon{N}$ exactly, by
averaging over all subsets of size $N$ (\cref{sec:supp-bon}), and average it over prompts, so the curves contain no
resampling noise. For the perfect picker this is the usual unbiased pass@$N$ estimator~\cite{chen2021codex}.

\paragraph{The exchange rate.}
Let $A_\pi(N)$ be the best-of-$N$ accuracy of \bstar with picker $\pi$, and let $a_{\mathrm{RL}}$ be the accuracy of
one image of \grpo. The exchange rate is
\begin{equation}
  \nstar_\pi = \min\{N \ge 1 : A_\pi(N) \ge a_{\mathrm{RL}}\}.
\end{equation}
We evaluate $A_\pi$ at every $N$ from 1 to 32 and interpolate linearly in $\log N$ where the curve crosses
$a_{\mathrm{RL}}$. If it does not reach $a_{\mathrm{RL}}$, we report $\nstar>32$. Unless we say otherwise, all intervals are 95\% percentile
intervals from a paired bootstrap over prompts, with the same resamples for all models and pickers. Differences are computed before rounding, so they can differ by
0.1 from the difference of two rounded numbers.

\paragraph{The registered test.}
We registered a single primary test. It asks whether the base model with the VLM picker and a budget of 16 samples
matches one sample of \grpo:
\begin{equation}
  D = \bon{16}(\bstar,\vlm) - \passk{1}(\grpo),
\end{equation}
on all 398 test prompts, judged by the registered rule, with a two-sided paired bootstrap over prompts at
$\alpha=0.05$. The budget of 16 was fixed in advance as a realistic serving budget. The registered plan named two more tests, both about the prompts that only one model solves
(\cref{sec:results-reach}), and the plan of each later human check named one test. All other $p$-values are
exploratory (\cref{sec:supp-stats}).

\paragraph{The first human check.}
\label{sec:method-human}
When the registered analysis was finished, the two automatic judges disagreed about $D$ (\cref{sec:results-real}). We
therefore asked people to decide, on every test prompt. For each prompt we took two images: the image that the VLM
picker chose among 16 random seeds of \bstar, and one random image of \grpo. Both images of a prompt went to the same
labeller, so that a strict or a generous labeller affects both models alike. We wrote down its analysis before most of its images were labelled,
but after the registered result, so we report it as exploratory (\cref{sec:supp-rules}). On the same 796 images we also compute $D$ under each automatic judge, which shows how each
judge differs from people.

\paragraph{Three human checks.}
\label{sec:method-checks}
People judged our comparisons in three checks. The first is the one just described. The second repeats it on a new
random draw of base and RL images, with a plan that was fixed before the images were drawn, and adds a picker of 38B
parameters (\cref{sec:supp-humanr}). The third is about the perfect picker. In its first part, people judge the image
that the perfect picker chooses among 8 and among 32 base images (\cref{sec:supp-humano}). In its second part, a
person looks through up to 8 base images of a prompt for a correct one (\cref{sec:supp-search}).
\Cref{tab:together} puts all three checks side by side.

\paragraph{Cost.}
\label{sec:method-cost}
We measure the A100-seconds per image spent on sampling ($g$), on decoding ($d$) and on the picker ($s_\pi$). One image
costs $g+d$, and best-of-$N$ costs $N(g+d+s_\pi)$ per prompt. The training of RL costs $H$ GPU-hours once. RL is
therefore the cheaper option after
\begin{equation}
  M^\star = 3600\,H \,/\, \big(N(g+d+s_\pi) - (g + d)\big)
\end{equation}
served prompts. We report it for $H=1{,}000$ GPU-hours.

\paragraph{Reach.}
\label{sec:method-reach}
An exchange rate assumes that the base model can match RL at some $N$. To test this, we count the prompts that only
one model solves within the images we drew (128 per model for the hard prompts). A single judge error can
create or hide such a prompt. So a person checks every prompt that the judges assign to one model only. The person goes through the images of that model which the judges passed, until one is confirmed. The person
also looks at the best images of the other model, in case the judges missed a correct one. The checked counts are compared with an exact McNemar test~\cite{mcnemar1947}. The prompts
that only \grpo solves then receive 896 more seeds from \bstar, 1,024 in total, and every pass of the base model among
them is checked by hand. Here ``unsolved'' always means that no confirmed pass was found among the images drawn.

%% file: sections/05_results.tex
\begin{figure*}[t]
\centering
\includegraphics[width=\textwidth]{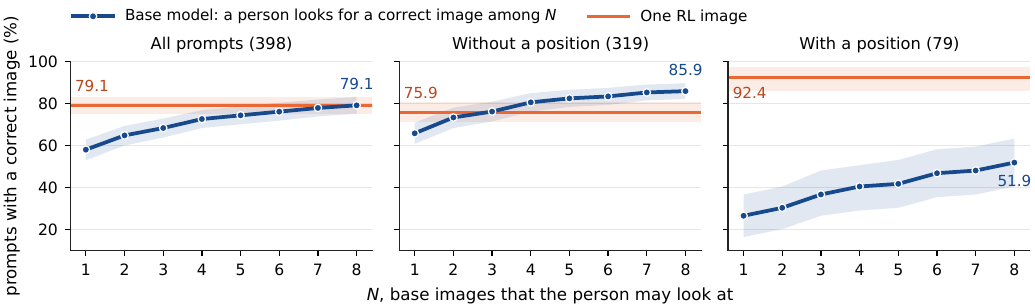}
\caption{\textbf{A person looks for a correct image among $N$ base images.} Blue: share of prompts for which at least
one of the first $N$ base images is correct, as judged by a person. Orange: share of prompts for which one RL image
is correct. Bands are 95\% bootstrap intervals of each share; \cref{tab:supp-search} gives the differences. Over
all prompts the base model reaches one RL image at about 8 images, where our series end. Without a position about 3
images are enough, and with a position 8 are far from enough. One person per prompt judged the base images one after another, which
may favour the base model somewhat (\cref{sec:supp-search}).}
\label{fig:person}
\end{figure*}

\section{Results}
\label{sec:results}

We present the results in the order in which a practitioner would ask for them. We start with the exchange rate
with a perfect picker (\cref{sec:results-rate}) and then use a picker that can be run in practice
(\cref{sec:results-real}). We show that the gap comes from one kind of prompt (\cref{sec:results-types}), count the
prompts that only one model solves (\cref{sec:results-reach}) and compare the costs (\cref{sec:results-cost}). The
last part looks at other models and another task (\cref{sec:results-other}).

\subsection{The exchange rate with a perfect picker}
\label{sec:results-rate}

\begin{table}[t]
\centering
\small
\setlength{\tabcolsep}{4.5pt}
\begin{tabular}{@{}lcccc@{}}
\toprule
 & Known & New & All & $\nstar$ \\
\midrule
\multicolumn{5}{@{}l}{\emph{One image}} \\
Base model & 27.2 & 67.1 & 41.2 & -- \\
RL model & 54.6 & 92.8 & 68.0 & -- \\
\midrule
\multicolumn{5}{@{}l}{\emph{Base model, best-of-16}} \\
PickScore & 31.7 & 73.9 & 46.5 & $>$32 \\
CLIPScore & 33.3 & 72.7 & 47.2 & $>$32 \\
GenEval reward & 43.2 & -- & -- & $>$32 \\
VLM picker & 46.0 & 82.3 & 58.8 & $>$32 \\
Perfect picker & 67.0 & 91.4 & 75.6 & 7.5 \\
\midrule
\multicolumn{5}{@{}l}{\emph{RL model, best-of-16}} \\
VLM picker & 68.6 & 94.9 & 77.8 & -- \\
\bottomrule
\end{tabular}
\caption{\textbf{Correct images (\%) on the test prompts} with known objects (258), new objects (140), and all 398.
An image is correct when both automatic judges pass it. $\nstar$: base images per prompt needed to match one RL image
($>$32: not reached). The GenEval reward cannot score new objects; its $\nstar$ is for known ones.}
\label{tab:main}
\end{table}

\begin{table*}[t]
\centering
\small
\setlength{\tabcolsep}{4pt}
\begin{tabular}{@{}lccccccc@{}}
\toprule
 & \multicolumn{3}{c}{All 398 prompts} & 156 hard prompts & 242 other prompts
 & \multicolumn{2}{c}{Judge passes (\%), best-of-16 / RL} \\
\cmidrule(lr){2-4}\cmidrule(lr){5-5}\cmidrule(lr){6-6}\cmidrule(l){7-8}
Judged by & Best-of-16 & RL & Difference & Difference & Difference & Correct (people) & Wrong (people) \\
\midrule
People & 70.4 & 75.9 & \m5.5 \ci{\m10.3}{\m0.8} & \m14.1 \ci{\m23.1}{\m5.1} & \phantom{\m}0.0 \ci{\m5.0}{\p5.0} & -- & -- \\
Both judges & 58.0 & 68.1 & \m10.1 \ci{\m14.8}{\m5.3} & \m18.6 \ci{\m27.6}{\m9.6} & \m4.5 \ci{\m9.9}{\p0.8} & 77.9 / 87.4 & 11.0 / \phantom{0}7.3 \\
Detector judge & 59.5 & 70.4 & \m10.8 \ci{\m15.6}{\m6.0} & \m21.2 \ci{\m30.1}{\m12.2} & \m4.1 \ci{\m9.5}{\p1.2} & 78.9 / 89.7 & 13.6 / \phantom{0}9.4 \\
VLM judge & 77.1 & 76.6 & \p0.5 \ci{\m3.3}{\p4.3} & \m1.9 \ci{\m10.3}{\p6.4} & \p2.1 \ci{\m0.8}{\p5.0} & 95.0 / 95.0 & 34.7 / 18.8 \\
\midrule
People, 2nd check & 71.1 & 77.6 & \m6.5 \ci{\m11.1}{\m2.0} & \m14.7 \ci{\m23.7}{\m5.8} & \m1.2 \ci{\m6.2}{\p3.7} & -- & -- \\
\quad 38B picker & 75.4 & 77.6 & \m2.3 \ci{\m7.0}{\p2.5} & \m8.3 \ci{\m17.9}{\p1.3} & \p1.7 \ci{\m2.9}{\p6.2} & -- & -- \\
\bottomrule
\end{tabular}
\caption{\textbf{Best-of-16 of the base model with the VLM picker against one RL image, judged by people and by the
automatic judges.} The first four rows use the same 796 images: for each prompt, the VLM picker's choice among one random set of 16 base images and one random RL image (\cref{tab:main} averages over all such sets). Columns 2--6: share of correct
images (\%) and the difference, best-of-16 minus RL, in points (negative: RL is better), with 95\% intervals.
``Both judges'': both must pass an image (the registered rule). Last two columns: of the images that people label correct (or wrong), the
share that the judge passes, for best-of-16 / RL; a fair judge gives two equal numbers. Last two rows: the second
human check, on a new random set of images, with the 8B picker and with a 38B picker.}
\label{tab:judges}
\end{table*}

One RL image is correct for 68.0\% of the test prompts, and one base image for 41.2\% (\cref{tab:main}). The base
model closes this gap of 27 points quickly when it may draw more images and a perfect picker selects among them
(\cref{fig:teaser}a). Its curve crosses the accuracy of the RL model at $\nstar=7.5$ (95\% interval 5.9 to 10.0). The
rate is 6.3 on the known-object prompts and 22.4 on the new-object prompts, where both models are already accurate and
the curve of the base model flattens early. The overall rate stays below ten under either judge alone: 6.3 under the
detector judge and 4.0 under the VLM judge. Nor does it depend on our choice of guidance: the base model at the
default guidance of 4.5 gives 7.2. So, by our judges, one RL sample is worth fewer than ten base samples with a
perfect picker.

People confirm this only for the prompts without a position. In the third human check, whose test was fixed in advance
(\cref{sec:supp-humano}), people judged the image that the perfect picker chooses. Over all prompts, its choice among
8 base images is 9.3 points behind one RL image (4.5 to 14.1). Without a position it is as good ($+0.3$ points,
$-4.4$ to $+5.0$). On position prompts it is 48.1 points behind, and still 40.5 points behind with 32 images. In this
check the judges underrate RL: they miss one in six correct RL images.

The perfect picker trusts the judges, so it can pass over a correct base image. In a second part of this third check,
the labellers therefore looked for a correct image themselves. When they failed the first base image of a prompt,
they saw the next ones, up to the eighth, until the first image they passed (\cref{sec:supp-search}). \Cref{fig:person} shows what they
found. A person finds a correct image among 8 base images for 79.1\% of the prompts. This is exactly the share of
RL images that people find correct in this check (a difference of 0.0 points, $-4.5$ to $+4.5$), while the choice of the judges among the same 8 images
is correct for 69.8\%. So when a person picks, the base model reaches one RL image over all prompts at about 8
images, where our series end. In 48\% of the bootstrap rounds, the curve does not reach the RL image within 8 images.
This average again mixes two cases. Without a position, about 3 base images are enough for a person (76.2\% with 3
images, against 75.9\% for one RL image). With a position, a person
finds a correct image among 8 for 51.9\% of the prompts, against 92.4\% for one RL image. In this part, one person per prompt
judged the base images of a prompt one after another, while each RL image was judged once, among images of other prompts. This
may favour the base model somewhat (\cref{sec:supp-search}). We therefore read the overall tie with care; the two
cases are clear.

By our judges, RL also stays ahead when both models may draw several images. Its lead in pass@$k$ falls from 26.8 points at $k=1$
to 11.8 at $k=32$, and the base model does not overtake it within this range.

\subsection{Best-of-16 with a practical picker}
\label{sec:results-real}

A perfect picker does not exist in practice, so we now use the VLM picker.

\begin{figure}[t]
\centering
\includegraphics[width=\linewidth]{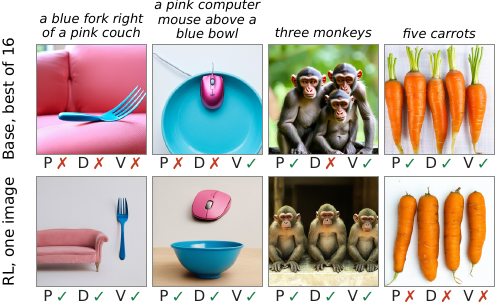}
\caption{\textbf{Four prompts from the first human check.} Top: the image that the VLM picker chose from 16 base
images. Bottom: one RL image. Under each image: verdicts of people (P), the detector judge (D) and the VLM judge (V).
Left to right: only RL is right; the VLM judge passes a wrong pick; the detector judge fails a correct pick; only
best-of-16 is right.}
\label{fig:examples}
\end{figure}

\paragraph{The registered test.}
With the VLM picker, best-of-16 of the base model is correct for 58.8\% of the prompts. This is 17.5 points above a
single base image, but 9.3 points below a single RL image: $D=-9.3$ points (95\% interval $-12.8$ to $-5.7$,
$p<0.001$). The registered test therefore rejects the hypothesis that best-of-16 matches RL. More images change
little: best-of-32 reaches only 60.3\%.

\paragraph{The picker falls short.}
Under the registered rule, other pickers leave a gap as well. The weaker
pickers do worse (\cref{tab:main}), and a picker with 38B instead of 8B parameters still leaves 8.4 points
(\cref{sec:supp-results}). Even the GenEval reward, on which RL was trained, is a worse picker than the VLM picker on
the known-object prompts (43.2\% against 46.0\%). Yet correct images are often there. Among 16 base images there is
a correct one for 75.6\% of the prompts, but the VLM picker finds it for only 58.8\%. So over all prompts, by our
judges, it is mainly the picker that keeps best-of-16 behind RL; on position prompts a correct image is also often
missing (\cref{sec:results-types}). The same picker also helps the RL model, which it raises from 68.0\% to
77.8\%.

\paragraph{The two judges disagree.}
The registered rule requires both judges to pass an image. Under the detector judge alone, the test gives almost the
same result as the rule, $D=-9.6$ points ($-13.3$ to $-6.0$). Under the VLM judge alone, it gives $D=+2.0$ points
($-0.7$ to $+4.8$) and finds no significant difference between best-of-16 and RL. So the conclusion of the registered
test depends on which judge one believes, and we asked people to decide.

\begin{table*}[t]
\centering\small
\setlength{\tabcolsep}{6pt}
\begin{tabular}{@{}llccc@{}}
\toprule
Human check & Who picks among $N$ base images & All prompts (398) & Without a position (319) & Position (79) \\
\midrule
First & 8B VLM picker, $N=16$ & \m5.5 \ci{\m10.3}{\m0.8} & \p2.2 \ci{\m2.8}{\p7.2} & \m36.7 \ci{\m48.1}{\m25.3} \\
Second & 8B VLM picker, $N=16$ & \m6.5 \ci{\m11.1}{\m2.0} & \p1.6 \ci{\m2.8}{\p6.0} & \m39.2 \ci{\m51.9}{\m27.8} \\
 & 38B VLM picker, $N=16$ & \m2.3 \ci{\m7.0}{\p2.5} & \p5.6 \ci{\p0.9}{\p10.3} & \m34.2 \ci{\m45.6}{\m22.8} \\
Third, first part & Perfect picker, $N=8$ & \m9.3 \ci{\m14.1}{\m4.5} & \p0.3 \ci{\m4.4}{\p5.0} & \m48.1 \ci{\m59.5}{\m36.7} \\
 & Perfect picker, $N=32$ & \m4.8 \ci{\m9.5}{0.0} & \p4.1 \ci{\m0.6}{\p9.1} & \m40.5 \ci{\m51.9}{\m29.1} \\
Third, second part & A person, $N=8$ & 0.0 \ci{\m4.5}{\p4.5} & \p10.0 \ci{\p6.0}{\p14.1} & \m40.5 \ci{\m51.9}{\m29.1} \\
\bottomrule
\end{tabular}
\caption{\textbf{All human checks side by side.} The picked base image minus one RL image, in points, as judged by
people (negative: RL is better), with 95\% intervals. In every row RL is far ahead on the position prompts, and no
row shows the base model behind on the prompts without a position. The first check (\cref{sec:results-real}) was
added after our main result; the second (\cref{sec:supp-humanr}) and the first part of the third
(\cref{sec:supp-humano}) were planned before their images were drawn or labelled. The second part of the third, in which a person looks through
the images, is weaker evidence (\cref{sec:supp-search}).}
\label{tab:together}
\end{table*}

\paragraph{The same test, judged by people.}
According to people (\cref{tab:judges}), the image that the VLM picker selects is correct for
70.4\% of the prompts and the RL image for 75.9\%. The difference is $-5.5$ points ($-10.3$ to $-0.8$, $p=0.03$).
People give the two images of a prompt different verdicts for 96 prompts, and in 59 of these the RL image is correct
(\cref{fig:examples} shows four prompts). The gap is not due to one labeller or to a check chosen after the result. A second check, labelled only by the two researchers and planned before its images were drawn, finds $-6.5$ points
($-11.1$ to $-2.0$, $p=0.006$) on a new random set of images (\cref{sec:supp-humanr}). Before these checks, the
registered plan itself had asked for a comparison by people on only 80 prompts. It is too small to show a gap of this size: there it is $-1.3$ points ($-8.8$ to $+6.3$,
\cref{sec:supp-results}).

So people confirm that best-of-16 does not catch up, but neither automatic judge measures the gap correctly.
On the same images, the registered rule reports $-10.1$ points and the detector judge $-10.8$, about twice the human
value. The VLM judge reports $+0.5$. Each differs from people by about five points, in opposite directions. Before a
correction for the many tests of this study, the difference from people is significant for the detector judge
($p=0.03$) and for the VLM judge ($p=0.006$), and nearly so for the registered rule ($p=0.05$). After a Holm
correction, only that of the VLM judge remains (\cref{sec:supp-stats}). \Cref{sec:analysis-judges} traces both errors to their source.

\paragraph{The gap comes from the hard prompts.}
\Cref{tab:judges} also splits the prompts. On the 156 hard prompts, people find best-of-16 correct for 37.8\% and RL
for 51.9\%, a gap of 14.1 points. On the other 242 prompts each is correct for 91.3\%. Because the hard prompts
were defined from the two compared models, we repeat the split with a definition that uses neither (the base model
at the default guidance passes the prompt at most 4 times out of 32). $D$ is then $-16.3$ points ($-25.5$ to
$-7.2$) on these 153 prompts and $+1.2$ points ($-4.1$ to $+6.5$) on the rest. So RL pays off on prompts that the base
model rarely solves; on the others the two are about equal. By prompt type, the gap comes from the prompts
that ask for a position, as we show next.

\subsection{The gap by prompt type}
\label{sec:results-types}

The numbers so far are averages over all prompts. This section shows that they mix two different cases.

\paragraph{Where RL helps.}
People find RL ahead by 37 points on the two prompt types that ask for a position, best-of-16 ahead by 18 points on
prompts with two counted classes, and the two are close on the remaining types (\cref{tab:supp-types}). On positions the
base model usually draws a wrong layout~\cite{gokhale2022visor,chatterjee2024spright}: even the image our judges pick
from 32 base images does not match one RL image, and 13 of the 20 prompts that only RL solves ask for a position
(\cref{sec:results-reach}). On the other
types a perfect picker needs only 3.9 base images (3.2 to 4.9, \cref{tab:supp-nstar-types}): the base model does draw
correct images, but the VLM picker often misses them.

\paragraph{All human checks side by side.}
\Cref{tab:together} collects every comparison that people judged. Two things hold in every row. On the 79 position
prompts, one RL image is far ahead of the base model, by 34 to 48 points. This does not depend on who picks among the
base images: the 8B or the 38B VLM picker with 16 images, the perfect picker with 8 or 32, or a person with 8. On the
319 prompts without a position, the base model is level with one RL image or ahead of it: the differences lie between
$+0.3$ and $+10.0$ points, and four of the six intervals include zero. The numbers for all prompts lie in between.
They depend on how many position prompts a test set contains, which is 19.8\% for ours and 18.1\% for GenEval.
Weighted to GenEval's mix of prompt types, the registered gap is 9.2 points instead of 9.3 (\cref{sec:supp-results}).

The rows are not independent repetitions. All of them use the same 398 prompts, images from the same 32 stored
seeds per prompt, and the same two labellers for nearly all labels. Nor do they carry the same weight as
evidence. The second check and the first part of the third named this split in their plans; the first check is where
we first saw it, and in the last row a person judged the base images of a prompt one after another, which may favour the
base model (\cref{sec:supp-search}). What they share
is the pattern. Of our conclusions, the one by prompt type therefore stands on the firmest ground. On position
prompts one RL image stays far ahead, whoever picks among the base images. On the other prompts 16 base images or
fewer are enough to match it.

\subsection{Prompts that only one model solves}
\label{sec:results-reach}

The exchange rate describes an average. It hides prompts that the base model may never solve, and we count them next.
Within up to 128 tries per model, both models solve 362 of the 398 prompts, and 13 are solved by neither. The RL
model alone solves 20 prompts, 17 of them with known objects, and the base model alone solves 3 ($p<0.001$;
\cref{tab:supp-coverage}). This count rests on the labels of one labeller. A second labeller judged the same images independently and agreed on 50 of the
55 prompts checked by hand ($\kappa=0.68$ over images). The five disputed prompts turned on how to
read a colour or a count. After the three labellers agreed on
written rules for these cases, the count is 20 against 5 ($p=0.004$). Under either reading, RL reaches further than
the base model.

How far out of reach are these 20 prompts? With 1,024 tries the base model solves 10 of them (exact 95\% interval 27\% to
73\%), among them all three with new objects. Eight of the ten that stay unsolved ask for a position and unusual
colours, such as \emph{a photo of a green laptop left of a pink pizza}. So for half of these prompts the base model can succeed, but only about once in several hundred images. For
the other half we found no correct base image. We checked by hand only the images that the judges passed, and the
judges miss some correct images, so the base model may solve more than half.

\subsection{Cost}
\label{sec:results-cost}

One image takes 1.75 seconds on an A100 for either model, because the RL weights are merged into the base model. The
VLM picker adds only 0.04 seconds per image, so the cost of best-of-16 is almost entirely the cost of sampling: 28.7
seconds per prompt, 16.4 times the cost of one RL image. RL has to be trained first. For the GenEval model, the owner of the Flow-GRPO code repository states in its
issue tracker that about 2,000 A800 GPU-hours are needed\footnote{\url{https://github.com/yifan123/flow_grpo/issues/22}};
we count them as A100 hours. By \cref{sec:method-cost}, each 1,000 GPU-hours of training is paid back after about 134,000 prompts served
with best-of-16, so 2,000 GPU-hours need about 267,000. A newer method reports a GenEval score of 0.95 for
SD3.5-M after about 90 H100 GPU-hours~\cite{choi2026rethinking}; even if each counts as three A100 hours, RL would
pay back after about 36,000 prompts. For a service with more prompts of the kind RL was trained on than the break-even number, RL
saves compute and gives more correct images. For fewer prompts, or for prompts without a reward, best-of-$N$
needs no training and gives most of RL's gain.

Stopping weak candidates early~\cite{sun2024speculative,yu2025dogtrim} lowers the cost of best-of-16, but no setting
closes the gap to RL (\cref{tab:supp-early}).

\subsection{Other models and another task}
\label{sec:results-other}

\paragraph{DiffusionNFT and text rendering.}
\nft behaves like \grpo: its exchange rate with a perfect picker is 5.2 (4.1 to 7.0), and with the VLM picker
best-of-32 does not reach it. On 200 prompts that ask for a rendered text, the OCR reward of \txt is a strong picker
that can be run, and with it the base model matches \txt at $N=5.6$ (4.5 to 7.2), judged by the VLM judge alone
(\cref{sec:supp-results}).

\ifflux
\paragraph{FLUX.1-dev.}
On the same \fluxPrompts{} test prompts, one image is correct for \fluxBaseOne\% of the prompts with the base
model and for \fluxRLOne\% with the adapter, and the exchange rate with a perfect picker is \fluxNstarPerfect, computed from $N=1,2,4,\dots,32$.
Best-of-16 with the VLM picker is correct for \fluxBonSel\%, so $D$ is \fluxD{} points;
the VLM judge alone gives \fluxDvlm{}, and no person checked these images (\cref{sec:supp-other}).
\fi

%% file: sections/06_analysis.tex
\section{Analysis}
\label{sec:analysis}

Two results need an explanation: why the judges disagree, and why RL helps most where the base model is weak.

\subsection{Why the judges disagree}
\label{sec:analysis-judges}

The human labels give the errors of each judge separately for the two models (\cref{tab:judges}, last two
columns). The detector judge is not equally sensitive to the two models. It passes 89.7\% of the correct RL images but
only 78.9\% of the correct best-of-16 images (\cref{fig:examples}, third column). Most of this difference comes from
the prompts with colours: about 21 points there, against about 5 on the others, in both human checks
(\cref{sec:supp-other}). For colours, our judge uses the same CLIP classifier as the GenEval reward, which RL was
trained to satisfy. But this shared part is not the only cause: with a colour classifier that was trained
elsewhere, the judge is still uneven on these prompts (\cref{sec:supp-judge2}). Nor is it because RL draws larger
objects, which a detector finds more easily: the RL model's objects are smaller on average (17\% of the image against
30\%). So a judge of the same kind
as the reward, a detector followed by a colour check of each object, favours RL and reports a larger gap than people
see.

The VLM judge is wrong in the other direction. It finds 95.0\% of the correct images of both models, but it also passes
34.7\% of the wrong best-of-16 images and only 18.8\% of the wrong RL images (\cref{fig:examples}, second column,
where the mouse lies in the bowl). The best-of-16 images are not random
images. A VLM picker preferred each of them to 15 others, so the wrong ones among them tend to be
wrong in ways that a VLM overlooks~\cite{panickssery2024self,koyama2026philautia,goel2025great}. A judge that is accurate on random
images therefore need not be accurate on selected images~\cite{yin2026ship}. These false passes hide the real gap.
Two other VLM judges, one of another family and one four times as large, pass the wrong picks in the same way
(\cref{sec:supp-judge2}).

The registered rule, which requires both judges, keeps the false passes low for both models (11.0\% and 7.3\%), but
it inherits the uneven sensitivity of the detector judge. So a comparison between a trained model and a search method
should use a judge of a different kind from both the reward and the picker, or check the selected images by hand.

\subsection{What RL changes, and when}
\label{sec:analysis-mech}

\begin{figure}[t]
\centering
\includegraphics[width=\linewidth]{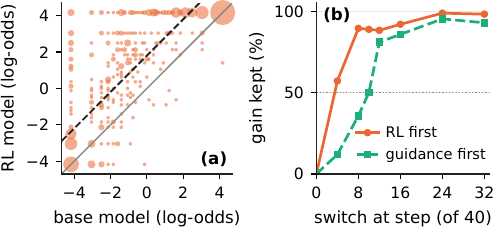}
\caption{\textbf{What RL changes, and when.} (a)~One point per group of test prompts with equal pass counts; larger
points hold more prompts. Axes: log-odds of a correct image, $\log\frac{k+0.5}{32-k+0.5}$ for $k$ passes in 32 seeds; the base model
uses guidance 4.5. On the grey line RL changes nothing; the dashed line adds 1.80 to every prompt. (b)~On 33 prompts with a large gain from RL, sampling
starts with one model and switches to the other at a given step (of 40). ``RL first'': share of the RL gain kept
when the base model finishes. ``Guidance first'': share of the guidance gain kept when guidance is switched off at
that step.}
\label{fig:mech}
\end{figure}

Since every seed gives the same starting noise to both models, we can follow single seeds. We compare the RL model
with its base model (guidance 4.5) on all $398\times32$ seeds.

\paragraph{No sign that RL mainly repairs near misses.}
The RL model turns 51.7\% of the failed base seeds into correct images and breaks 7.7\% of the successful ones. One
might expect that the rescued seeds are mostly those where the base image was almost right. We see no sign of this:
the score that one judge gives to a failed base image does not predict whether the other judge passes the RL image
from the same seed (AUROC 0.50, 0.47 to 0.52, comparing seeds within a prompt).

\paragraph{The gain differs strongly between prompts.}
If RL only raised the weight of the base images that the reward likes by a fixed factor, the odds of a correct image
would rise by a similar factor for every prompt (\cref{sec:supp-tilt}). On average the log-odds rise by 1.80, a factor
of six in the odds (\cref{fig:mech}a), but the prompts scatter widely around this average: sampling noise cannot
explain 69\% of the variation between prompts. A single shift predicts an accuracy of 61.5\% for the RL model, below
the observed 68.0\%. The gain is larger where the base model is weaker, as on the hard prompts
(\cref{sec:results-real}).

\paragraph{The outcome is decided early.}
To see when RL acts, we let one model begin the sampling and the other finish it, for 33 prompts with a large gain from RL
(\cref{fig:mech}b). If the RL model takes the first 8 of the 40 steps and the base model the remaining 32, 89\% of the
gain of RL remains. In the reverse order only 30\% remains. Half of the gain is already reached after 3.5 steps, at a
noise level of 0.97. Guidance acts later: when it is switched off after 8 steps, 36\% of its gain remains, and it
needs 10 steps for half. On these prompts, RL thus acts mostly in the first steps of sampling, where the coarse structure of an image
is set. This fits the two observations above: we see no sign that RL mainly repairs near misses, and it helps most where
the base model rarely draws a correct arrangement.

%% file: sections/07_limitations.tex
\section{Discussion and limitations}
\label{sec:limitations}

This section asks what could change these results, and says what the study does not cover.

\paragraph{Better pickers and search.}
By our judges a perfect picker closes the gap, and a larger picker may narrow it. Averaged over all sets of 16 base images, the registered rule finds gaps of
6.2 and 8.4 points with VLM pickers of 14B and 38B parameters. In the second check, people and the rule both find the
38B picks 4.3 points better than the 8B picks, and people find $D=-2.3$ points ($-7.0$ to
$+2.5$, \cref{sec:supp-humanr}). Averaged over all such sets, the rule finds the 38B picks under one point better than the 8B
picks, so part of this gain may be chance.
Pickers of two other families, Gemma-3~\cite{gemmateam2025gemma3} and the original VQAScore model, leave 11.4 and
13.9 points. Two searches at the cost of best-of-16, over 32 early previews or near the
noise of the best image so far~\cite{ma2025inference}, gain only 1.8 points each. A picker that asks separate questions about objects, counts and positions~\cite{fan2026keystone} may do better
than our single question. Our VLM judge is such a picker, and with it best-of-16 stays 6.4 points behind under the
detector judge (\cref{sec:supp-results}). Other search
methods~\cite{singhal2025fk,jain2025dts} may use the samples better, but they too need a score that tells correct
from wrong.

\paragraph{Other detector judges.}
On the 258 known-object prompts, the official GenEval code and the GenEval reward on which RL was trained show the
same uneven sensitivity in both human checks. They pass 56--70\% of the correct best-of-16 images but 81--86\% of the
correct RL images, and report gaps of 10 to 17 points where people see 6 to 7 (\cref{sec:supp-other}).

\paragraph{Scope.}
\ifflux
Our main results are for one base model, SD3.5-M, with a check of the main test on FLUX.1-dev,
\else
Our results are for one base model, SD3.5-M,
\fi
and for public checkpoints, so we cannot vary the amount of RL training. A second Flow-GRPO run leaves a similar gap
(10.9 points); a run without the KL penalty, which is less accurate on our prompts, leaves none (\cref{sec:supp-other}). Our prompts
follow the GenEval templates used to train RL, which may favour RL. Our judges measure whether an image matches its prompt, not its quality, which RL is known to
affect~\cite{wang2026grpoguard}; on PickScore and on an aesthetic score~\cite{schuhmann2022laion5b}, the best-of-16
and RL images are about equal (\cref{sec:supp-results,tab:supp-quality}). RL does make the images of a prompt
more alike, even the correct ones: 32 images count as 3.1 clearly different ones for RL and 4.3 for the base model
(\cref{sec:supp-results}). Two researchers made nearly all labels of the human checks, and most images have one label. We tested only with
the first labeller whether a person can guess the model behind an image (\cref{sec:supp-prompts}). We see no new risk
to society, but best-of-16 needs about 16 times the energy.

%% file: sections/08_conclusion.tex
\section{Conclusion}
\label{sec:conclusion}

We asked how many base model samples one RL sample is worth. The answer differs by prompt. Without a position, a few
base samples are enough: a perfect picker needs about four by our judges, a person who looks through the
images needs about three, and best-of-16 with an off-the-shelf VLM picker is about as good as RL. There, much of what RL adds is within
reach of the base model, and the limit of best-of-$N$ is the picker. On prompts that ask for a position, RL is far
ahead: a person finds a correct image among eight base samples for only half of the prompts, where one RL image is
correct for nine in ten. All our human checks show these two cases. Over all prompts, best-of-16 with the VLM picker stays 5.5 and 6.5 points behind one RL
image in two human checks. Both kinds of automatic judge that we tested misjudge this gap: our detector judge makes it larger and our
VLM judge removes it, so comparisons between trained models and search methods should check the selected images by
hand.

%% file: sections/09_llm_statement.tex
\section*{LLM Usage Statement}
\label{sec:llm-statement}

We used a large language model, Claude (Anthropic), through the Claude Code tool. It wrote most of the code for
analysis, experiments, and plotting, which the author
reviewed. Together with the author it drafted the analysis plan that we registered. It checked our references against their sources.
It paraphrased the text of this paper. It did not label any
image. When labellers had general questions about the labelling rules, it helped to answer them, and each answer was
added to the written rules of all labellers. The author made the scientific decisions, reviewed the code, checked
the numbers and the citations against their sources, and takes full responsibility for the content of the paper.

%% file: sections/A_supplement.tex
\clearpage
\setcounter{page}{1}\renewcommand{\thepage}{S\arabic{page}}
\maketitlesupplementary\raggedbottom
\appendix
\setcounter{table}{0}\renewcommand{\thetable}{S\arabic{table}}\renewcommand{\theHtable}{S\arabic{table}}
\setcounter{figure}{0}\renewcommand{\thefigure}{S\arabic{figure}}\renewcommand{\theHfigure}{S\arabic{figure}}

This supplement gives the details behind the paper and follows its order. We first list what we fixed in advance
and where we departed from it (\cref{sec:supp-rules}). We then describe the prompts, the judges and the labelling
(\cref{sec:supp-prompts}), and explain how we compute best-of-$N$ and the exchange rate
(\cref{sec:supp-bon,sec:supp-rate}). Further results follow (\cref{sec:supp-results}). We then give the theory
behind the per-prompt analysis, more related work, statistical details and more judges
(\cref{sec:supp-tilt,sec:supp-related,sec:supp-stats,sec:supp-judge2}). The last section reports the second part of
the human check of the perfect picker, in which a person looks through several base images of a prompt
(\cref{sec:supp-search}). All
differences are computed before rounding, so they can differ by 0.1 from the difference of two rounded numbers.

\section{The registered plan and departures from it}
\label[appendix]{sec:supp-rules}

The main test and the choices behind it were fixed in a plan before we saw any test image. This section shows the
plan and every place where we departed from it, so that the reader can tell planned results from later ones. The
plan was committed to a private repository with a time stamp before the first test image was generated. \Cref{tab:supp-rules} lists every rule of the plan that makes a choice, and
the choice it made.

\begin{table}[t]
\centering\small
\setlength{\tabcolsep}{3pt}
\begin{tabular}{@{}L{0.24\linewidth}L{0.72\linewidth}@{}}
\toprule
Rule & What was fixed, and the outcome \\
\midrule
Guidance of \bstar & The guidance scale in $\{1, 2.5, 4.5, 7\}$ that gives the highest best-of-16 accuracy on dev with the VLM picker; ties go to the higher pass@1. Chosen: 7 (\cref{tab:supp-dev}). \\
Budget & $N=16$ for the primary test. \\
VLM picker & InternVL3.5-8B with the question given in \cref{sec:setup}. \\
Primary test & $D$ on all 398 test prompts under the registered rule (both judges pass), two-sided, $\alpha=0.05$. \\
Hard prompts & \grpo or \bstar passes at most 4 of 32 seeds. All 156 such prompts received seeds 32--127. \\
Early exit & Stop all images at a step in $\{4,8,12,16\}$ and finish the best $m\in\{1,2,4\}$. The rule takes the cheapest pair (step, $m$) that is within 2 points of full best-of-16 on dev. Chosen: step 8, $m=1$. \\
Deep sampling & The prompts that only \grpo solves after the hand check (at most 25) get 1,024 seeds of \bstar.
Registered test: do more than half of them stay unsolved? Ten of 20 do, which is not more than half ($p=0.59$,
one-sided exact binomial test). \\
Switching models & Prompts of the analysis subset on which \grpo beats \base by at least 0.3 on seeds 8--31. Selected: 33. \\
Detector judge & Two thresholds tuned once on 240 labelled dev images, then kept fixed: 0.15 for presence and 0.25 for counting. \\
Counts & A count level is kept only if both judges count it reliably (\cref{sec:supp-prompts}). Kept: up to five. \\
\bottomrule
\end{tabular}
\caption{\textbf{The rules of the registered plan.} Each row is a rule that makes a choice. The right column says
what the rule fixed and, where it had to choose, what it chose. Dev: the 87 dev prompts. ``Switching models'': the
prompts on which we let one model begin the sampling and the other finish it (\cref{sec:analysis-mech}).
Analysis subset: 86 test prompts drawn at random in advance.}
\label{tab:supp-rules}
\end{table}

\paragraph{Departures from the plan.}
We list every point at which the study did not follow the plan, or added to it.
\begin{enumerate}
  \setlength\itemsep{1pt}
  \item \emph{The first human check} (\cref{sec:method-human}) was not in the plan. We decided on it after
  the registered analysis, when the two judges disagreed. Its analysis was written down before labellers~2 and~3 made
  their first label; 160 of its images reuse labels that labeller~1 had made a day earlier, with the general rules
  only (\cref{sec:supp-prompts}). We report it as exploratory.
  \item \emph{A second human check} (\cref{sec:supp-humanr}) was added after the registered analysis, with its own
  plan committed before any of its images was drawn. After finishing the labels, labeller~2 reported a possible mistake near one
  image. The seven images around it were labelled again in a separate set. The planned analysis keeps the first
  labels.
  \item \emph{A third human check} (\cref{sec:supp-humano}), of the perfect picker, was added later, with its own plan
  committed before any of its tasks or labels. In its second part, a labeller who failed the first base image of a prompt was shown the next
  ones. On the day of the check, the labellers had time for only 26 of these images. They finished the other 801 six
  days later. This was one day after the
  last date that the plan had set for the analysis, so we report it apart from the first part, as a late addition
  (\cref{sec:supp-search}).
  \item \emph{A human check of FLUX.1-dev was planned but not labelled.} Its plan and the images to be labelled
  were fixed before any label, in the same way as for the second human check. No image of it has been labelled yet, so FLUX.1-dev has no human labels (\cref{sec:supp-other}).
  \item \emph{The registered early-exit setting worked less well on the test prompts than on dev.} Compared with full
  best-of-16, it lost 1.2 points on dev and 7.2 points on test (\cref{tab:supp-early}).
  \item \emph{A third labeller and written rules.} The plan had two labellers. Labellers~1 and~2 disagreed on five
  prompts of the reach analysis (the prompts that only one model solves). A third labeller judged these prompts, and
  all three then labelled the disputed images again under written rules (\cref{sec:supp-prompts}). The registered
  result uses the first-round labels of labeller~1. We report the result under the written rules beside it. It is not
  an independent measurement, because the labellers knew that these images were disputed.
  \item \emph{Labeller~1 judged cut fruit inconsistently in the first round.} They passed some cut watermelons as
  pink and failed others. This is the main source of the disagreement on the five disputed prompts, and the written
  rules settle it.
  \item \emph{One prompt was missing from the first deep-sampling run.} A prompt that only \grpo solves
  (\emph{a photo of four cakes and three chairs}) was left out of the first list by mistake. Its 896 extra images
  were generated later on a second A100 of the same type. Before using them, we generated 128 earlier images again on
  this card, and all 128 were identical to the originals in every pixel.
  \item \emph{Two confirming labels were added late.} For two prompts in the reach analysis, the passes of \grpo were at
  first taken from the judges alone. The hand checks that confirm them were done a day later, and both passes held.
  \item \emph{Self-agreement after one day.} The plan asked the first labeller to relabel 100 images after at least
  three days. This was first done after only one day (96 of 100 equal, $\kappa=0.90$). A second
  set of 100 images was then relabelled as planned, three and a half days after the first labels, with almost the
  same result (96 of 100 equal, $\kappa=0.91$).
  \item \emph{Sampling of hard prompts.} The plan allowed us to use a random subset of the hard prompts if there were
  more than 300. There were 156, so all were used and no weighting is needed.
  \item \emph{Who labelled.} The plan expected that the people who analyse the data would also label the images. In
  the end, three people who are not authors made all labels (\cref{sec:supp-prompts}).
  \item \emph{An LLM assistant helped to answer general questions from the labellers} about the rules (see the LLM usage
  statement). It never saw a question about a specific image, and every answer was added to the written rules that
  all labellers received.
  \item \emph{FLUX.1-dev on all test prompts.} The plan named 200 of the test prompts and 16 base seeds for
  FLUX.1-dev. Before the first FLUX image was drawn, we fixed a rule to use all 398 prompts and 32 base seeds if they
  were finished by a date fixed in advance (\cref{sec:supp-other}). They were. On the 200 registered prompts, $D$ is $-11.1$ points
  ($-16.9$ to $-5.5$).
  \item \emph{Tests outside the plan.} The registered plan asked for tests only of $D$, the reach count and the deep
  sampling, and the plans of the later human checks each added one test. All other $p$-values in the paper and here
  are exploratory. We show in \cref{sec:supp-stats} which of them remain after a correction for many tests.
\end{enumerate}

\paragraph{Sampler settings.}
All SD3.5-M models use fp16 weights with an fp32 VAE, prompts cut to 128 T5 tokens and an empty negative
prompt, as in the evaluations of the Flow-GRPO and DiffusionNFT authors. All SD3.5-M images were generated on NVIDIA
A100-SXM4-40GB GPUs, on which a given seed gives the same image on every card.
\ifflux
FLUX.1-dev is sampled with 28 steps at $512\times512$, as in the evaluation of its adapter~\cite{chen2025mapreducelora}.
It uses the default guidance of FLUX.1-dev (3.5), bf16 weights, an fp32 VAE and prompts cut to 128 T5 tokens. We
kept this guidance for both FLUX models and did not tune it for best-of-$N$ as we did for SD3.5-M, so this comparison may be less favourable to
the base model than the main one. The FLUX images were generated on two \fluxGPU{}
GPUs, and both FLUX models ran on the same cards.
\fi

\paragraph{Names.}
The registered plan and our code refer to prompt sets, judges and stages by short codes.
\Cref{tab:supp-names} maps them to the names used in this paper.

\begin{table}[t]
\centering\small
\setlength{\tabcolsep}{3pt}
\begin{tabular}{@{}L{0.52\linewidth}L{0.44\linewidth}@{}}
\toprule
In this paper & In the plan and code \\
\midrule
GenEval prompts & P1 \\
Known-object prompts & P2 \\
New-object prompts & P3 \\
Text prompts & P5 \\
Analysis subset & mechanism subset \\
VLM judge, detector judge & J1, J2 \\
Correct (both judges pass) & V \\
GenEval reward & V0 \\
VLM picker & Sel \\
Perfect picker & oracle \\
\base, \bstar, \bzero & B, B* (B7), B0 \\
\grpo, \nft, \txt & G, N, T \\
First 32 seeds; hard prompts & Stage A; Stage B \\
Reproduction check; judge check & Gate 0; Gate 1 \\
\bottomrule
\end{tabular}
\caption{\textbf{Names in this paper and in the registered plan.} In the main paper, ``the base model'' is \bstar
and ``the RL model'' is \grpo, except in \cref{sec:analysis-mech} and \cref{fig:mech}, which use \base.}
\label{tab:supp-names}
\end{table}

\section{Prompts, judges and labelling}
\label[appendix]{sec:supp-prompts}

This section gives the details behind \cref{sec:setup}: how we built the prompts, how we checked the two judges,
and how people labelled the images.

\paragraph{Prompts.}
The known-object prompts have five types: a count of five (\emph{a photo of five teddy bears}), two counted classes
(\emph{a photo of three books and four bottles}), three objects, three colour bindings (\emph{a photo of a red
microwave, a brown broccoli, and an orange tennis racket}), and a colour with a position (\emph{a photo of a green
laptop left of a pink pizza}). The new-object prompts use the six types of GenEval (one object, two objects,
counting, colours, position and colour attribution) with objects outside COCO, for example \emph{a photo of a crown
right of a sunflower}. Of 70 candidate objects, 12 have fewer than five LVIS photos, and 6 are found by a judge in
fewer than 80\% of their photos. We kept the other 52. Sixteen new-object prompts use an object that is a sub-type of
a COCO class. Leaving them out changes the results little (\cref{tab:supp-ablations}).

Of the 200 text prompts, 50 are taken from the test set of Flow-GRPO. The other 150 ask for new texts, 50 in each of
three ranges of length (18--29, 30--45 and 46--70 characters; the longest text has 53). None of the 50 prompts from
Flow-GRPO is among its training prompts, but 19 of their target texts also occur in a training prompt. We keep them because they show the task as its authors test it, and we note
that they are not fully held out.

\paragraph{The count check.}
Larger counts are hard for automatic judges, so the plan kept a count level only if both judges count it reliably.
For each level (5, 6 and 7), 50 dev images were counted by hand with the target number hidden. We called a judge
reliable if three conditions hold: the lower 80\% confidence bound of
the exact-count accuracy is at least 0.85, the sensitivity is at least 0.8, and the specificity is at least 0.9. Only
the count of five passed, and each judge counted 47 of the 50 images correctly.

\paragraph{The detector judge.}
The detector judge has two score thresholds, one for presence and one for counting. Both were chosen on 240
hand-labelled dev images to maximise the balanced accuracy. At the chosen
thresholds the judge passes 84\% of the correct images and rejects 96\% of the wrong ones. The boxes come from
OWLv2 (base model, patch size 16, ensemble weights). As in GenEval, the colour of each detected object is judged by
a zero-shot CLIP ViT-L/14 classifier with GenEval's three prompt templates. This colour classifier is the same as
in the GenEval reward; the detector is not, and it took no part in training.

\paragraph{Who labelled.}
Three people labelled images. None of them is an author, and none took part in the design of the study or knew its
hypotheses. Labeller~1 labelled first. Their labels cover the checks of the registered plan, the 80 prompts labelled
first and the reach analysis (\cref{tab:supp-labels}). Labellers~2 and~3 are researchers who work on vision-language
models. Each labeller worked alone and in their own session. They did not see each other's labels, the verdicts of
the judges or any result. All three took part voluntarily and agreed to the use of their labels in this study. We recruited no one from outside, asked the labellers for no personal information, and showed them only images
that the models had generated, most of everyday objects and a few of people. So we judged that the study needed no ethics review, and we did not seek one.

\paragraph{The labelling screen.}
An image is shown with a short list of statements taken from its prompt, for example ``there is a laptop'', ``the
laptop is green'', ``the laptop is left of the pizza''. The labeller marks each statement as true or false, and the
image is correct only if all are true. The model behind the image is not shown, and the images come in random order. There are two exceptions, and the plans foresaw both. In the second part of the third human check, a labeller saw
several images of one prompt in a row (\cref{sec:supp-search}). In the first round of the reach check, labeller~1 judged the images group by group, one group for
each prompt and model, and the groups of the prompts that only RL solves came first. The model was still hidden, but
the order could hint at it. Labeller~2 judged the same images in random order. Of the 280 base images, labeller~1
failed 23 that labeller~2 passed and passed 2 that labeller~2 failed. Of the 127 RL images, labeller~1 failed 21 that
labeller~2 passed and passed none that labeller~2 failed. So labeller~1 was the stricter of the two for both models,
and if anything more so for RL. We see no sign that the order favoured RL. In the count
check only the name of the object was shown, never the target number.

\paragraph{The written rules.}
All labellers received the following general rules.
\begin{itemize}
  \setlength\itemsep{1pt}
  \item An object counts if an ordinary viewer, without reading the prompt, would call it that, even if it is placed
  oddly.
  \item A colour counts only if the object itself has it, not its background, holder or surroundings.
  \item ``A right of B'' (or left, above, below) means that A is beside B on that side, not in front of or inside B.
  \item If one statement clearly fails, the image fails.
\end{itemize}
In the first round, the labellers had read some colour and counting cases differently. So we wrote the
following rules and gave them to all three.
\begin{itemize}
  \setlength\itemsep{1pt}
  \item \emph{Objects with two colours.} An object has a colour if that colour covers most of its visible surface. For a cut fruit,
  judge the flesh when it fills most of the visible fruit; shades from pink to light red count as pink.
  \item \emph{Holders.} The colour must belong to the object, not to its plate, pan or stand. A pizza is black only if
  black covers most of the pizza itself.
  \item \emph{Traffic lights.} The colour of a traffic light is the colour of its housing, not of a lit lamp.
  \item \emph{Counting.} Count every separate, recognisable object of the kind, including partly hidden ones. Pictures,
  reflections and prints of objects do not count. ``Exactly $N$'' fails with one too many or one too few. Two fused
  objects count as two only if each is still recognisable as a whole.
  \item \emph{Potted plants.} A potted plant is a living plant in soil in a container; cut flowers in a vase are not.
\end{itemize}
Labellers~2 and~3 labelled the first human check (\cref{sec:method-human}) with all of these rules. Labeller~1 had
labelled 80 of its prompts earlier, with the general rules only.

\begin{table}[t]
\centering\small
\setlength{\tabcolsep}{3pt}
\begin{tabular}{@{}L{0.64\linewidth}rl@{}}
\toprule
What was labelled & Labels & Labeller \\
\midrule
Count check & 150 & 1 \\
Thresholds of the detector judge & 240 & 1 \\
50 images that the judges passed & 50 & 1 \\
First 80 prompts, plus 80 reward picks & 320 & 1 \\
\quad the same, 100 of these images & 100 & 2 \\
\quad the same, 100 images after one day & 100 & 1 \\
\quad the same, another 100 after 3.5 days & 100 & 1 \\
Reach: prompts one model solves & 407 & 1 \\
\quad the same prompts & 415 & 2 \\
\quad disputed images & 70 & 3 \\
\quad colour rules, 70 images & 210 & 1, 2 and 3 \\
\quad counting rules, 42 images & 126 & 1, 2 and 3 \\
Deep sampling: base passes, RL images & 181 & 1 \\
Text rendering: check of the text judge & 60 & 1 \\
First human check & 736 & 2 and 3 \\
Second human check & 1,196 & 2 and 3 \\
\quad seven images after a reported mistake & 7 & 2 \\
Human check of the perfect picker & 1,076 & 2 and 3 \\
\quad second part & 827 & 2 and 3 \\
\midrule
Total & 6,371 & \\
\bottomrule
\end{tabular}
\caption{\textbf{All human labels, by task and labeller.} Each row is one labelling task, and indented rows belong to
the task above them. ``Deep sampling'': the passes of the base model in 1,024 tries, with RL images hidden among them. The first human check also uses 160 of the 320 labels of the first 80 prompts. This gives 796 images, of which 100 were
labelled twice. The second human check has 1,063 images, of which 133 were labelled twice.}
\label{tab:supp-labels}
\end{table}

\paragraph{Agreement.}
\Cref{tab:supp-agree} gives every agreement number. The lowest agreement is on the reach images in the first round.
There, labellers~1 and~2 disagree on 46 images, and on 44 of them labeller~1 failed the image and labeller~2 passed
it. Most of these are colour cases that the written rules later settled. For whole prompts, the two labellers reached
the same decision for 50 of 55.

\begin{table}[t]
\centering\small
\setlength{\tabcolsep}{3pt}
\begin{tabular}{@{}L{0.56\linewidth}ccc@{}}
\toprule
Comparison & Images & Equal & $\kappa$ \\
\midrule
Labellers 2 and 3, first check & 100 & 98 & 0.95 \\
Labellers 2 and 3, second check & 133 & 130 & 0.94 \\
Labellers 2 and 3, perfect picker & 123 & 122 & 0.98 \\
Labellers 1 and 2, first 80 prompts & 100 & 93 & 0.84 \\
Labeller 1, again after one day & 100 & 96 & 0.90 \\
Labeller 1, again after 3.5 days & 100 & 96 & 0.91 \\
Labellers 1 and 2, reach, first round & 407 & 361 & 0.68 \\
\midrule
\multicolumn{4}{@{}l}{\emph{After the written rules (not independent)}} \\
Labellers 1 and 2, reach & 408 & 396 & 0.92 \\
All three, colour cases & 70 & 70 & 1.00 \\
All three, counting cases & 42 & 42 & 1.00 \\
\bottomrule
\end{tabular}
\caption{\textbf{Agreement between labellers.} Images: the number of images that both sides labelled. Equal:
how many of them received the same label. $\kappa$: Cohen's kappa on correct / not correct. ``Labeller 1, again'': labeller~1 labelled the same images a
second time. The rows below the line come after the
written rules. They are not independent, because the labellers knew that these images were disputed.}
\label{tab:supp-agree}
\end{table}

Hiding the model behind an image helps only if people cannot tell the two models apart by their look. To test
this, we showed 48 images to labeller~1, half from each model, and asked them to guess for each image whether it came
from the RL model or the base model. They guessed right for 20, which is no
better than chance.

\section{The best-of-\texorpdfstring{$N$}{N} estimator}
\label[appendix]{sec:supp-bon}

The best-of-$N$ curves in the paper are exact expected accuracies, not the outcome of one random draw. This section
shows how we compute them. For a single prompt we have $n$ images with picker scores $s_i$ and verdicts
$e_i\in\{0,1\}$. Best-of-$N$ draws $N$ of
them at random, keeps the one with the highest score, and is correct if that image is correct. Its expected accuracy
can be computed exactly. We sort the images by score, highest first. The image at rank $j$ is kept exactly when it
is among the $N$ drawn images and none of the $j-1$ images above it is. This happens with probability
$\binom{n-j}{N-1}/\binom{n}{N}$, so
\begin{equation}
  \bon{N} = \sum_{j=1}^{n-N+1} \frac{\binom{n-j}{N-1}}{\binom{n}{N}}\; \bar e_j ,
\end{equation}
where $\bar e_j$ is the verdict at rank $j$, averaged over the images with the same score. Giving every image in a
tied group the mean verdict of the group is the same as breaking ties at random. With the judge as the picker (the
perfect picker), this is the usual unbiased pass@$N$ estimator~\cite{chen2021codex}. Early exit with $m=1$ uses the
same formula, with the picker scores of the early previews in place of the final scores. For $m>1$ we average over 4,000 random draws of $N$ seeds per prompt.

\section{Details of the exchange-rate analysis}
\label[appendix]{sec:supp-rate}

This section gives three details of the analysis in \cref{sec:method}: how we find the exchange rate, whether
best-of-$N$ overfits the picker, and how we check by hand the prompts that only one model solves.

\paragraph{Crossing point.}
We evaluate the mean best-of-$N$ curve at every integer $N$ from 1 to 32. The exchange rate $\nstar$ is the point
where this curve reaches the accuracy of the RL model. We find it by interpolating linearly in $\log N$ between the
two values of $N$ on either side of the crossing. In each bootstrap resample, we recompute both the curve
and the accuracy of the RL model on the same resampled prompts, so the interval for $\nstar$ includes the
uncertainty of both. If the curve never reaches the accuracy of the RL model in a resample, that resample gives
$\nstar=\infty$. If this happens in more than 2.5\% of the resamples, we report the upper end of the interval as
``$>32$''.

\paragraph{Does best-of-$N$ overfit the picker?}
The judged accuracy of best-of-$N$ can peak and then fall as $N$ grows, because a larger search also finds more of
the images that fool the picker~\cite{huang2025bestofn,khalaf2025hacking}. We see little of this within our budgets.
With the VLM picker, the curve over all 398 prompts still rises at $N=32$. On the hard prompts it rises up to about
$N=120$ and then stays flat (0.04 points lower at $N=128$). The only
exception is the new-object prompts: there the curve peaks at $N=20$ and is 0.6 points lower at $N=32$.

\paragraph{The hand check of prompts that only one model solves.}
For every prompt that the judges assign to one model only, a person checks two things. First, they go through the
passes of the model that solves it, one by one, until one pass is confirmed. Second, they look at the ten images of
the other model with the highest picker scores, in case the judges missed a pass there. We check in the same way the prompts that the judges
assign to both models but on which one model has at most four passes, because a single judge error can hide a prompt
of this kind.

\section{Further results}
\label[appendix]{sec:supp-results}

This section adds to \cref{sec:results,sec:analysis}. We first show the choices and checks behind the main
comparison and how the gap varies over prompt types. We then turn to cost, to the prompts that only one model
solves, and to further checks. Three subsections report a second human check, a human check of the perfect picker,
and other checkpoints, searches and judges.\paragraph{One RL image, several accuracies.} The accuracy of one RL image takes several values in the paper and here. The model is always the same; the judge and the images differ. Under the registered rule, one RL image is correct for 68.0\% of the prompts over all 32 seeds (\cref{tab:main}) and for 68.5\% over seeds 0--15 (\cref{tab:supp-ckpt}). The single random RL image of the first and of the second human check is correct for 68.1\% and 67.1\% (\cref{tab:judges,tab:supp-humanr}). People pass more of the same images, 75.9\% and 77.6\%, and 79.1\% of the new RL image of the third check (\cref{tab:supp-humano}). The difference comes mostly from correct RL images that the judges miss (\cref{sec:analysis-judges}). Best-of-16 behaves in the same way: 58.8\% under the rule averaged over all sets of 16 base images, 58.0\% and 57.0\% on the single random sets of the two checks, and 70.4\% and 71.1\% by people.

\paragraph{The choice of the base model.}
\Cref{tab:supp-dev} shows the dev results behind the choice of guidance. The plan chose by best-of-16 with the VLM
picker, which selects a guidance of 7. Choosing by pass@32 would have selected 2.5, which is better with a perfect
picker at large $N$ and worse with the VLM picker. On the test prompts, the default guidance of 4.5 gives almost the
same result as 7 (\cref{tab:supp-ablations}).

\begin{table}[t]
\centering\small
\begin{tabular}{@{}lccc@{}}
\toprule
Guidance & One image & Best of 16, VLM picker & pass@32 \\
\midrule
1 & 22.1 & 41.6 & 59.8 \\
2.5 & 39.6 & 54.8 & \textbf{78.2} \\
4.5 & 43.6 & 57.0 & 77.0 \\
7 & \textbf{45.3} & \textbf{57.5} & 75.9 \\
\bottomrule
\end{tabular}
\caption{\textbf{Guidance scale of the base model on the 87 dev prompts.} Share of prompts with a correct image (\%)
under the registered rule (both judges pass): one image, best-of-16 with the VLM picker, and pass@32 (best-of-32
with a perfect picker). Bold: the best value in each column. The plan takes the guidance with the best value of
best-of-16, which is 7.}
\label{tab:supp-dev}
\end{table}

\paragraph{Reproduction of published scores.}
Before the main study, we checked that our setup reproduces the published GenEval scores (\cref{tab:supp-gate0}).
One detail matters here. In the GenEval reward code that Flow-GRPO uses for training (the file
\path{reward_server/gen_eval.py} of its reward server), the detector receives the images as arrays in RGB channel
order, and the detection library reads such arrays as BGR. The official GenEval code gives the detector the image
files, which the library loads in BGR order. We report the score computed in both ways. The two differ by 0.03 for
\base and by at most 0.002 for the two RL models.

\begin{table}[t]
\centering\small
\begin{tabular}{@{}lccc@{}}
\toprule
 & Published & As in the RL code & Official code \\
\midrule
\base & 0.63~\cite{liu2025flowgrpo} & 0.651 & 0.680 \\
\grpo & 0.95~\cite{liu2025flowgrpo} & 0.933 & 0.935 \\
\nft  & 0.94~\cite{zheng2026diffusionnft} & 0.930 & 0.931 \\
\bottomrule
\end{tabular}
\caption{\textbf{Reproduction of the published GenEval scores} on the 553 GenEval prompts, seeds 0--3. ``As in the
RL code'': scored as in the GenEval reward code used to train Flow-GRPO, where the detector receives arrays in RGB
channel order. ``Official
code'': the released GenEval code. Scored as in the RL code, all three models are within 0.03 of the published score.}
\label{tab:supp-gate0}
\end{table}

\paragraph{Variants of the main comparison.}
Does the main result depend on our choices? \Cref{tab:supp-ablations} changes one part of the registered comparison
at a time. The exchange rate with a perfect
picker stays between 4 and 8 on all prompts under every judge and guidance. The difference $D$ (best-of-16 with the
VLM picker minus one RL image) stays between $-8.6$ and $-15.0$ points in every row except the VLM judge alone.

\begin{table}[t]
\centering\small
\setlength{\tabcolsep}{3pt}
\begin{tabular}{@{}L{0.43\linewidth}cc@{}}
\toprule
Variant & $D$ (points) & $\nstar$ \\
\midrule
Registered comparison & \m9.3 \ci{\m12.8}{\m5.7} & 7.5 \ci{5.9}{10.0} \\
Base model at guidance 4.5 & \m9.6 \ci{\m13.0}{\m6.1} & 7.2 \ci{5.8}{9.2} \\
Detector judge alone & \m9.6 \ci{\m13.3}{\m6.0} & 6.3 \ci{5.0}{8.1} \\
VLM judge alone & \p2.0 \ci{\m0.7}{\p4.8} & 4.0 \ci{3.2}{5.3} \\
Known objects only & \m8.6 \ci{\m13.5}{\m3.8} & 6.3 \ci{4.8}{8.7} \\
\quad GenEval reward as judge & \m15.0 \ci{\m19.5}{\m10.4} & -- \\
New objects only & \m10.5 \ci{\m15.0}{\m6.4} & 22.4 \ci{10.5}{$>$32} \\
\quad without COCO sub-types & \m11.9 \ci{\m17.0}{\m7.3} & 23.6 \ci{10.4}{$>$32} \\
Hard prompts, 128 samples & -- & 12.2 \ci{8.4}{17.5} \\
\bottomrule
\end{tabular}
\caption{\textbf{Variants of the main comparison.} Each row changes one part of the registered comparison. $D$:
best-of-16 of the base model with the VLM picker minus one RL image, in points (negative: RL is better). $\nstar$:
base images per prompt that a perfect picker needs to match one RL image. Small numbers are 95\% intervals. Only the
first row is the registered test. Under the VLM judge alone, the VLM picker reaches the RL model at $\nstar=9.8$ (6.2
to 20.2). In all other rows it does not reach the RL model within the budget.}
\label{tab:supp-ablations}
\end{table}

\paragraph{The gap by prompt type.}
The registered test averages over eleven prompt types, and \cref{tab:supp-types} shows them one by one. We did not
plan this split, so it is exploratory here. The two later human checks, of best-of-16 and of the perfect picker, named the split by position
in their plans, and both find the same: RL is far ahead on positions and about level elsewhere
(\cref{sec:supp-humanr,sec:supp-humano}). The gap differs a lot between types. On the two types that ask for a position,
one RL image is ahead of best-of-16 by 34 and 39 points under the registered rule, and people see about the same (37
points on both). On prompts with two counted classes the order is reversed. The RL model is correct for only 16.1\%
of these prompts and best-of-16 for 35.4\%, and people confirm a lead of 18.3 points for best-of-16. The training
prompts of \grpo ask for a count of one class only. This may be the reason, but we did not test it. For the two types
with colour bindings (three colour bindings and colour attribution), the registered rule reports a gap of about 21 points, but people see little of it (6.7 and 0.0
points). On these two types the rule is also much stricter than people, for both models. For example, on the best-of-16 images of the first human check it passes 28\%
and 57\%, where people pass 72\% and 90\%. This fits the uneven sensitivity of the detector
judge (\cref{sec:analysis-judges}). On the easy types both options are almost always correct.

So the average gap depends on the mix of prompt types, which we fixed in the registered plan. Our mix is close to
that of the official GenEval benchmark: 19.8\% of our test prompts ask for a position, and 18.1\% of GenEval's (100
of 553). If we weight the types to match GenEval's six categories, the gap is $-9.2$ points ($-11.8$ to $-6.7$)
under the registered rule and $-5.5$ points ($-9.2$ to $-1.7$) under people, the same as on our mix. With every type
weighted equally, it is $-10.7$ points ($-13.7$ to $-7.6$). Leaving out any one type from our mix keeps the registered gap between $-4.0$ and
$-14.3$ points. Without the two position types, best-of-16 roughly matches RL: $-2.2$ points ($-5.7$ to $+1.3$)
under the registered rule and $+2.2$ points ($-2.8$ to $+7.2$) under people. If we resample whole types instead of
single prompts, the 95\% interval widens to $-23.3$ to $+3.9$ points. This interval answers a different question,
the gap for a randomly chosen type, and with eleven types it is only a rough check.

\paragraph{Why positions differ.}
On the position types, what the base model mainly lacks is correct images, not a good picker. With a perfect picker,
best-of-32 of the base model is correct for 58.3\% of the colour-and-position prompts and 84.2\% of the position
prompts, while one RL image is correct for 73.4\% and 92.4\%. On every other type, a perfect picker over 16 base
images reaches or passes one RL image. For three colour bindings, for example, it reaches 66.1\% against 46.0\% for
RL, and it is the VLM picker that falls short (24.7\%). The prompts that only RL solves tell the same story: 13 of
the 20 ask for a position, and so do 8 of the 10 that the base model still misses after 1,024 tries
(\cref{tab:supp-reach}). The exchange rate with a perfect picker splits in the same way
(\cref{tab:supp-nstar-types}). On the 319 prompts of the other types, the base model matches one RL image at
$\nstar=3.9$ (3.2 to 4.9), well below the 7.5 of all prompts. On the 79 position prompts it does not reach one RL
image within 32 samples. Single types without a position can still have a high rate, such as counting with new
objects (15.3). This happens when one RL image is already correct for most prompts, so that the curve of the base
model flattens near the top.

\begin{table*}[t]
\centering\small
\setlength{\tabcolsep}{5pt}
\begin{tabular}{@{}lc ccc c cc c@{}}
\toprule
 & & \multicolumn{4}{c}{Registered rule, all images} & \multicolumn{3}{c}{People, one image each} \\
\cmidrule(lr){3-6}\cmidrule(l){7-9}
Prompt type & Prompts & One base & Best-of-16 & One RL & $D$ (points) & Best-of-16 & One RL & $D$ (points) \\
\midrule
\multicolumn{9}{@{}l}{\emph{Known objects}} \\
Count of five & 18 & 12.0 & 38.8 & 54.0 & \m15.2 \ci{\m28.8}{\m1.9} & 38.9 & 38.9 & 0.0 \ci{\m33.3}{\p33.3} \\
Two counted classes & 60 & 9.9 & 35.4 & 16.1 & \p19.3 \ci{\p12.1}{\p26.8} & 36.7 & 18.3 & \p18.3 \ci{\p5.0}{\p31.7} \\
Three objects & 60 & 74.3 & 91.5 & 82.9 & \p8.7 \ci{\p0.9}{\p16.4} & 85.0 & 85.0 & 0.0 \ci{\m13.3}{\p13.3} \\
Three colour bindings & 60 & 15.1 & 24.7 & 46.0 & \m21.3 \ci{\m30.6}{\m12.0} & 71.7 & 78.3 & \m6.7 \ci{\m20.0}{\p6.7} \\
Colour and position & 60 & 14.2 & 34.6 & 73.4 & \m38.9 \ci{\m48.0}{\m29.8} & 48.3 & 85.0 & \m36.7 \ci{\m50.0}{\m23.3} \\
\midrule
\multicolumn{9}{@{}l}{\emph{New objects}} \\
One object & 23 & 99.5 & 100.0 & 100.0 & 0.0 \ci{0.0}{0.0} & 100.0 & 95.7 & \p4.3 \ci{0.0}{\p13.0} \\
Two objects & 25 & 95.0 & 99.6 & 99.0 & \p0.6 \ci{0.0}{\p1.7} & 100.0 & 100.0 & 0.0 \ci{0.0}{0.0} \\
Counting & 26 & 48.1 & 79.5 & 91.8 & \m12.3 \ci{\m23.0}{\m3.3} & 96.2 & 100.0 & \m3.8 \ci{\m11.5}{0.0} \\
Colours & 26 & 86.9 & 93.2 & 95.9 & \m2.7 \ci{\m7.6}{\p1.0} & 100.0 & 100.0 & 0.0 \ci{0.0}{0.0} \\
Position & 19 & 22.0 & 58.6 & 92.4 & \m33.8 \ci{\m52.5}{\m14.1} & 52.6 & 89.5 & \m36.8 \ci{\m63.2}{\m10.5} \\
Colour attribution & 21 & 37.9 & 53.5 & 75.3 & \m21.8 \ci{\m36.3}{\m9.1} & 90.5 & 90.5 & 0.0 \ci{\m14.3}{\p14.3} \\
\bottomrule
\end{tabular}
\caption{\textbf{The gap by prompt type.} Share of prompts with a correct image (\%) for one base image, best-of-16 of
the base model with the VLM picker, and one RL image. $D$: best-of-16 minus one RL image, in points (negative: RL is
better), with 95\% intervals. The left part uses the registered rule on all images. The right part uses the labels of
people on the two images per prompt of the first human check. RL is far ahead on the two types that ask for a
position, and best-of-16 is ahead on prompts with two counted classes. The split by type was not planned.}
\label{tab:supp-types}
\end{table*}

\paragraph{Several RL samples.}
So far we compared many base samples with a single RL sample. What if RL also gets several samples? With a perfect
picker, $M$ samples of the RL model are matched by 7.5 base samples for $M=1$ and by 15.6 for $M=2$.
For $M=4$ the base model does not reach the RL model within 32 samples. So, between $M=1$ and $M=2$, the exchange rate grows
roughly in proportion to $M$ (slope 1.05 on a log-log scale, 95\% interval 0.83 to 1.34). With only these two values
inside our budget, this is a rough check: the second RL sample is worth about as many base samples as the first.

\begin{figure}[t]
\centering
\includegraphics[width=\linewidth]{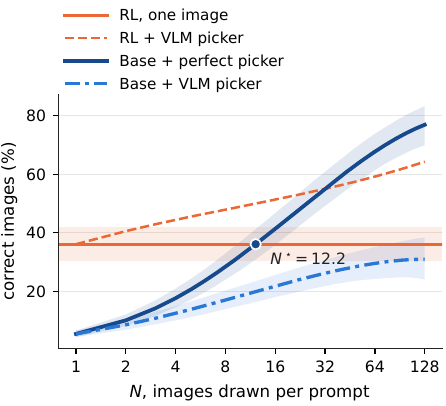}
\caption{\textbf{Best-of-$N$ of the base model on the 156 hard prompts, up to $N=128$.} Share of prompts with a
correct image under the registered rule. With a perfect picker, the base model passes one RL image at about 12
samples. With the VLM picker, it stays below one RL image at every $N$. Bands are 95\% intervals.}
\label{fig:supp-hard}
\end{figure}

\paragraph{Hard prompts with 128 samples.}
A prompt is hard if either model passes it at most 4 times in 32 seeds. Of the 156 hard prompts, 79 are hard only
for the base model, 70 for both models and 7 only for \grpo. So which prompts count as hard depends on the two models that we compare.
This is why the main text repeats the split into hard and other prompts with a definition that uses only the base
model at guidance 4.5. Most hard
prompts ask for two counted classes (47), a colour and a position (39) or three colour bindings (34), and 48 of the
79 position prompts are hard. \Cref{fig:supp-hard} shows the curves on the 156 hard prompts up to $N=128$. One RL
image is correct for 36.1\% of these prompts and one base image for 5.4\%. With a perfect picker, the base model
reaches the RL model at $\nstar=12.2$ and rises to 76.9\% at $N=128$. With the VLM picker it reaches 30.9\% at
$N=128$ and stays below one RL image. The registered crossover check asks whether the base model ever overtakes RL
in pass@$k$, the share of prompts with a correct image among $k$. With 95\% bands that hold for all $k$ at once, RL
stays ahead at every $k$ by our judges: by 11.8 points at $k=32$ on all prompts (7.4 to 16.3), and by 14.1 points at
$k=128$ on the hard prompts (5.1 to 23.1). These curves describe the hard prompts only and are not combined with the 32-seed curves of the other
prompts.

\paragraph{Early exit.}
Best-of-16 spends most of its cost on images that the picker then discards. Early exit saves part of this cost. It
stops all 16 images at an early step, lets the picker score a preview of each, and finishes only the best $m$. \Cref{tab:supp-early} gives all twelve settings on dev and on test. The registered plan picked the cheapest
setting within 2 points of full best-of-16 on dev, which was step 8 with $m=1$. On the 87 dev prompts several
settings look as good as full best-of-16. On the 398 test prompts, however, the loss shrinks steadily as the step
gets later and $m$ gets larger. This suggests that the dev set was too small for this choice. Stopping at step 16
with $m=2$ matches full best-of-16 at 53\% of its cost, but we chose this setting after seeing the test results.

The same previews allow a simple search over more candidates at the same cost. We start all 32 seeds, score their
previews at step 16 and finish the best 4. This costs about as much as 15 full images, close to best-of-16. It is
correct for 60.6\% of the prompts: 1.8 points more than best-of-16 and 7.5 points less than one RL image. On the
position prompts it reaches 44.3\%, against 78.0\% for RL. More candidates do not help much when the picker cannot
find the correct ones, and on positions there are few correct ones to find.

\begin{table}[t]
\centering\small
\setlength{\tabcolsep}{4pt}
\begin{tabular}{@{}cccccc@{}}
\toprule
 & & \multicolumn{2}{c}{Correct (\%)} & & \\
\cmidrule(lr){3-4}
Step & $m$ & Dev & Test & Seconds & Cost \\
\midrule
4 & 1 & 49.5 & 47.0 & \phantom{0}6.3 & 22\% \\
4 & 2 & 53.6 & 51.4 & \phantom{0}7.9 & 28\% \\
4 & 4 & 55.3 & 54.2 & 11.1 & 39\% \\
\textbf{8} & \textbf{1} & 56.3 & 51.5 & \phantom{0}8.9 & 31\% \\
8 & 2 & 58.2 & 54.3 & 10.3 & 36\% \\
8 & 4 & 58.0 & 56.3 & 13.2 & 46\% \\
12 & 1 & 56.4 & 55.9 & 11.4 & 40\% \\
12 & 2 & 56.2 & 57.2 & 12.7 & 44\% \\
12 & 4 & 56.3 & 57.9 & 15.2 & 53\% \\
16 & 1 & 57.6 & 57.1 & 14.0 & 49\% \\
16 & 2 & 57.6 & 59.0 & 15.1 & 53\% \\
16 & 4 & 59.2 & 59.0 & 17.3 & 60\% \\
\midrule
\multicolumn{2}{@{}l}{Full best-of-16} & 57.5 & 58.8 & 28.7 & 100\% \\
\bottomrule
\end{tabular}
\caption{\textbf{Early-exit best-of-16 with the VLM picker.} All 16 images are stopped at the given step (of 40), the
picker scores a preview of each, and the best $m$ are finished. Correct: share of prompts with a correct image under
the registered rule, on the 87 dev prompts (Dev) and the 398 test prompts (Test). Seconds: time per prompt on one
A100. Cost: share of the time of full best-of-16. The registered setting is in bold.}
\label{tab:supp-early}
\end{table}

\paragraph{Cost and break-even.}
The break-even point in \cref{sec:results-cost} rests on the following timings. On one A100, sampling one image
takes 1.71 seconds for \bstar and \grpo and 0.85 seconds for \nft and \bzero, which
need one model call per step instead of two. Decoding takes 0.04 seconds, the VLM picker 0.04, PickScore 0.01, the VLM
judge 0.16 and the detector judge 0.09 seconds per image. The text encoders run once per prompt, and we leave them
out.
\Cref{tab:supp-breakeven} gives the number of served prompts after which RL is cheaper, for a range of training costs.

\begin{table}[t]
\centering\small
\begin{tabular}{@{}rrr@{}}
\toprule
Training cost & \multicolumn{2}{c}{Prompts until RL is cheaper than} \\
\cmidrule(l){2-3}
(GPU-hours) & best-of-16 & best-of-32 \\
\midrule
100 & 13,400 & 6,500 \\
500 & 66,800 & 32,400 \\
1,000 & 133,700 & 64,700 \\
2,000 & 267,300 & 129,500 \\
4,000 & 534,700 & 258,900 \\
\bottomrule
\end{tabular}
\caption{\textbf{Break-even number of served prompts.} After this many prompts, training RL and then drawing one RL
image per prompt costs less than best-of-16 or best-of-32 of the base model with the VLM picker.}
\label{tab:supp-breakeven}
\end{table}

\paragraph{The prompts that only one model solves.}
\Cref{tab:supp-coverage} gives the counts of \cref{sec:results-reach} separately for the two prompt sets.

\begin{table}[t]
\centering\small
\setlength{\tabcolsep}{4pt}
\begin{tabular}{@{}lccccc@{}}
\toprule
Prompts & RL only & Base only & Both & Neither & $p$ \\
\midrule
Known objects & 17 & 3 & 226 & 12 & 0.003 \\
New objects & \phantom{0}3 & 0 & 136 & \phantom{0}1 & 0.25 \\
All & 20 & 3 & 362 & 13 & $<$0.001 \\
\bottomrule
\end{tabular}
\caption{\textbf{Who solves a prompt at all?} Number of test prompts on which a model has at least one correct
image among up to 128 tries. Every prompt in the columns ``RL only'' and ``Base only'' was
checked by hand. $p$: exact McNemar test that compares the columns ``RL only'' and ``Base only''.}
\label{tab:supp-coverage}
\end{table}

\begin{figure}[t]
\centering
\includegraphics[width=\linewidth]{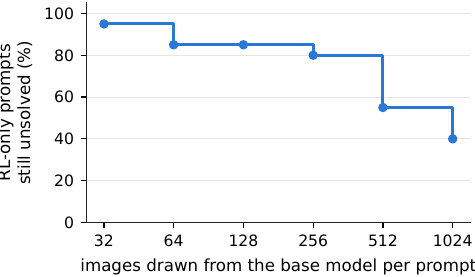}
\caption{\textbf{More tries for the base model on the 20 prompts that only RL solved.} The curve shows the share of
these prompts for which no base image has yet passed both judges, as the number of tries grows. It falls to 40\% at
1,024 tries. After a hand check of these passes, 50\% remain
unsolved. The curve counts the passes of the judges, so it is below 100\% already at 128 tries: for 3 of the 20
prompts, the judges pass a base image within 128 tries, but the hand check found these images wrong.}
\label{fig:supp-reach}
\end{figure}

\Cref{tab:supp-reach} lists the 20 prompts of the registered reach analysis and whether 1,024 tries of the base model
solve them. \Cref{fig:supp-reach} shows how the number of unsolved prompts falls with the number of tries according to
the judges. For two of the prompts that the judges count as solved at 1,024 tries, the hand check found no correct base image, which leaves 10 of the 20 solved.
The three prompts that only the base model solves are one prompt with two counted classes (\emph{a photo of four skateboards and two
baseball bats}) and two prompts with three colour bindings (for example \emph{a photo of a yellow banana, a brown
broccoli, and a white knife}). Under the written rules there are five such prompts. With these rules, one prompt
also leaves the list of 20 and another enters it, and the base model again solves 10 of the 20. If we count
only the prompts that both labeller~1 and labeller~2 confirm, the base model solves 8 of 18. If we count every prompt
that either of them confirms, it solves 12 of 23.

\begin{table}[t]
\centering\small
\setlength{\tabcolsep}{3pt}
\begin{tabular}{@{}L{0.86\linewidth}c@{}}
\toprule
Prompt (each begins with ``a photo of'') & Solved \\
\midrule
five teddy bears & yes \\
four handbags and four bowls & yes \\
three books and four bottles & yes \\
four cakes and three chairs & yes \\
a brown dining table above a purple microwave & yes \\
a green stop sign below a brown backpack & yes \\
a white person left of a brown toilet & yes \\
a white globe and a pink watermelon & yes \\
a crown right of a sunflower & yes \\
a pillow right of a pretzel & yes \\
\midrule
three oranges and four apples & no \\
a red microwave, a brown broccoli, and an orange tennis racket & no \\
a brown broccoli right of a pink bowl & no \\
a blue fork right of a pink couch & no \\
a white spoon right of an orange train & no \\
a green laptop left of a pink pizza & no \\
a black oven right of a blue banana & no \\
a purple couch below a yellow orange & no \\
a black pizza right of a blue traffic light & no \\
a green bicycle above a red orange & no \\
\bottomrule
\end{tabular}
\caption{\textbf{The 20 prompts that only the RL model solves in 128 tries}, and whether the base model solves them in
1,024 tries (Solved: a pass of the base model confirmed by hand). The known-object prompts with colours combine
colour words and COCO classes at random, in the GenEval templates. One COCO class is ``person'', so a few prompts pair
a person with a colour word; the written rules treat this case like any other.}
\label{tab:supp-reach}
\end{table}

\paragraph{The 80 prompts labelled first.}
These are the human labels that the registered plan asked for. The plan sized them only to confirm the direction of a
gap of 15 points or more, not as a test. Before the first human check, labeller~1 labelled three images for each of 80 test prompts (40 with known and 40
with new objects): a random image of \grpo, a random image of \bstar, and the best-of-16 pick of the VLM picker. The
pick of the GenEval reward was labelled separately, for 80 known-object prompts, most of them not among these 80. Labeller~1 finds 76.3\% of the RL images correct,
57.5\% of the random base images, 75.0\% of the picks of the VLM picker and 63.8\% of the picks of the GenEval reward.
On these 80 prompts the gap between best-of-16 and RL is $-1.3$ points ($-8.8$ to $+6.3$). This is consistent with the
$-5.5$ points on all 398 prompts, but it is too imprecise to show such a gap on its own. An image that both judges
pass is almost always correct (precision 1.00 for RL images, 0.97 for random base images and 0.98 for the picks of
the VLM picker). What differs is the share of correct images that the judges find (recall): 90\% for RL, 78\% for
random base images and 83\% for the picks.

\paragraph{DiffusionNFT.}
We repeat the main comparison with the second RL model. \nft needs no guidance, so one of its images costs 0.90
seconds with decoding, about half of the 1.75 seconds of
the base model. One
image of \nft is correct for 63.5\% of the test prompts (50.7\% on known and 87.1\% on new objects). Against
\bstar, the exchange rate with a perfect picker is 5.2 (4.1 to 7.0). It is 5.7 under the VLM judge alone and 4.5
under the detector judge alone. With the VLM picker, best-of-16 of \bstar is 4.7 points below one image of \nft, and
the curve does not reach \nft within 32 samples (the lower end of the interval is 26.3). We also compare with \bzero,
which costs the same per image as \nft. Here best-of-16 with a perfect picker reaches 53.3\%, and the exchange rate
is above 32 (lower end 24.6).

\paragraph{Text rendering.}
On the 200 text prompts, \txt renders the requested text correctly in 57.5\% of its images and the base model in
32.8\%. The text judge is the VLM judge, which reads the text without seeing the target. Checked on 60
hand-labelled images, it has a precision of 0.81 and a recall of 0.95. With a perfect picker, the base model reaches
\txt at 3.3 samples (2.8 to 3.8). Unlike in the main task, here a strong picker can be run: the OCR reward, which reads the
image and compares it with the requested text. With this picker the base model reaches \txt at 5.6 samples (4.5 to
7.2). The VLM picker, which is not built to read text, does not reach \txt within 32 samples (the lower end of the
interval is 10.3). So when the reward is a good picker, best-of-$N$ is a real alternative to RL. These results rest on the text judge alone, without a human check.

\paragraph{Further checks.}
\emph{Resolution.} Sampling the base model at $1024\times1024$ instead of $512\times512$ changes its accuracy by $+1.5$
points ($-2.6$ to $+5.5$) on the analysis subset, 86 test prompts drawn at random in advance, so the low resolution puts the base
model at no clear disadvantage.
\emph{Seeds.} A seed that is good for one prompt is usually not good for other prompts. The seed explains 0.1\% of the
variance of the verdicts for \grpo and 0.2\% for \bstar. The four seeds that are best for the base model on training prompts give
43.1\% on the test prompts against 41.0\% for the other seeds.
\emph{Object size.} The boxes that the detector finds cover 17\% of the image for \grpo and 30\% for \bstar. Both
judges pass images with larger objects more often. From the fifth of images with the smallest objects to the fifth
with the largest, the pass rate rises from 53\% to 76\% for the VLM judge and from 47\% to 63\% for the detector
judge. Since the RL objects are smaller, the judges do not favour the RL model through the size of its objects.
\emph{Reward and judge.} On the known-object prompts the log-odds of a correct image rise by 2.23 from \base to \grpo
under the GenEval reward, computed as in the RL code, and by 1.90 under our judges. Here every prompt counts equally; \cref{tab:supp-tilt} weights each prompt by how precisely its change is measured, which gives 1.67 under our judges. So about 85\% of the gain on the reward carries
over to judges that the model was not trained on.
\emph{Preference scores.} Besides the aesthetic score of \cref{tab:supp-quality}, PickScore and CLIPScore give a
rough signal of image quality that neither the judges nor the VLM picker use. The mean PickScore is 23.31 for one RL image, 23.06 for one
base image, and 23.32 for the image that the VLM picker selects among 16 base images. So the RL image is 0.25 above
the base image (0.21 to 0.30) and level with the best-of-16 image (difference 0.01, $-0.04$ to $+0.06$). For CLIPScore
the three values are 32.33, 30.95 and 31.63. Both scores also reward the match between image and prompt, so they do
not isolate image quality, and they say nothing about diversity. They only show that two common scores do not find
the RL images worse.
\emph{Image quality.} To measure the look of an image without its prompt, we use the LAION aesthetic
score~\cite{schuhmann2022laion5b}. It is a small model on CLIP ViT-L/14 image features~\cite{radford2021clip} that
predicts how people rate the look of an image, on a scale from 1 to 10. For every test prompt, we scored the images
of both human checks, plus one more random base image and one more random RL image (\cref{tab:supp-quality}). On SD3.5-M, the image that best-of-16 delivers, a random base image and an RL image
all score about 5.4, and no difference is larger than 0.04 points; every interval includes zero. On FLUX.1-dev the
images of the RL adapter score 0.06 points higher than the base images (0.03 to 0.08), which is small. So by this
score neither RL nor best-of-16 costs image quality, and the VLM picker does not trade looks for correctness.
\paragraph{A stronger picker.} A better picker would narrow the gap, and we tried three ways to get one. First, larger
models of the same family. We asked the same question to InternVL3.5 with 14B and 38B parameters instead of 8B, on
the same stored images. Under the registered rule, best-of-16 reaches 58.8\%, 61.8\% and 59.6\% with the 8B, 14B and
38B pickers. The gap to one RL image (68.0\%) is then $-9.3$, $-6.2$ ($-9.7$ to $-2.8$) and $-8.4$ points ($-11.9$ to
$-4.9$). So under the registered rule the 38B picker is hardly better than the 8B picker (difference $+0.8$ points, $-0.8$ to $+2.6$), although people found its picks better on the one random set of images that they labelled (\cref{sec:supp-humanr}). The
detector judge alone gives the same picture ($-9.6$, $-6.7$ and $-8.6$ points). Only the VLM judge, which is of the
same kind as the pickers, favours the larger pickers ($+2.0$, $+4.0$ and $+4.9$ points).
Why does size help so little? For each prompt, we measured how well a picker ranks the 32 base images, using the area
under the ROC curve (AUROC) with the verdict of the registered rule as the label. Averaged over prompts, it is 0.815,
0.828 and 0.833 for the 8B, 14B and 38B pickers. All three agree well with the rule on the position types (0.87 to
0.89) and poorly on the two types with colour bindings (0.66 to 0.69, at every size). On colour bindings, however,
the rule itself is much stricter than people (\cref{tab:supp-types}), so these low values may reflect the rule as
much as the pickers. The first human check labelled only the picks of the 8B picker. The larger pickers choose a
different image for about 70\% of the prompts, so most of their picks have no human labels in that check. The larger pickers
ran on another machine. To check that machine, we scored 256 images on it again with the 8B picker, and the ranks agree with
the stored scores (Spearman correlation above 0.999). On this machine, a single A100, the 38B picker takes 0.18 seconds
per image and the 8B picker 0.06. With 0.18 seconds in place of the 0.04 seconds of the 8B picker in our main
timings, best-of-16 with the 38B picker costs about 30.9 seconds per prompt instead of 28.7, about 8\% more,
because sampling the images still dominates the cost.

Second, pickers of other families. InternVL3.5 uses a Qwen3 language model, like the VLM judge. So we also asked the
same question to Gemma-3-12B, which shares no part with the judge. We also used the original VQAScore model,
CLIP-FlanT5-XXL, which was trained for this kind of question~\cite{lin2024vqascore}. Both are worse pickers than
InternVL3.5-8B: best-of-16 reaches 56.6\% and 54.1\%, which leaves gaps of $-11.4$ ($-15.0$ to $-7.7$) and $-13.9$
points ($-17.5$ to $-10.3$). They also rank the images of a prompt less well (AUROC 0.78 and 0.74), and they too are
poor on colour bindings (0.59 to 0.66). Under the VLM judge alone, the picks of these two pickers do worse than those
of the 8B picker ($-1.2$ and $-2.5$ against $+2.0$ points). This agrees with \cref{sec:analysis-judges}: a VLM judge
most easily overlooks the mistakes of a picker that is similar to it.

Third, we let two VLMs of different families pick the image together. We added the score of the VLM judge to the
score of the VLM picker, after standardising each within a prompt. We then judged with the detector judge alone,
because it is the only judge that is not part of this picker. Best-of-16 rises from 60.8\% to 64.7\%, and the gap to
one RL image (70.4\%) falls from 9.6 to 5.7 points (2.0 to 9.4). The VLM judge alone, used as a picker, does about as well. Its score
combines the answers to separate questions about each object, count, colour and position (\cref{sec:setup}), and
its picks reach 64.0\%, a gap of 6.4 points (2.7 to 10.1) under the detector judge. With either picker the gap
narrows but does not close. The
detector judge favours the RL model (\cref{sec:analysis-judges}), so the remaining gap is probably smaller than 5.7
points.
Adding PickScore, or PickScore and CLIPScore, to the VLM picker makes it worse ($D=-17.1$ and $-17.4$ points under
the registered rule).
\paragraph{Variety of the images.} RL is known to make the images of one prompt more alike~\cite{tan2026pec,liu2026drift},
and we measured how much this happens on our images. We describe each image by its DINOv2
feature~\cite{oquab2023dinov2}, and for each prompt we compute the Vendi score~\cite{friedman2023vendi} of its 32
images. It can be read as the number of clearly different images, between 1 and 32. The base model at guidance 7
gives 4.28 such images per prompt and the RL model 3.08 (difference $-1.20$, $-1.32$ to $-1.07$). The RL images are
less varied for 85\% of the prompts. The base model at guidance 4.5, from which RL started, gives 4.86. The loss of
variety is not only a loss of wrong images. On the 202 prompts where both models have at least eight correct images,
eight correct RL images count as 2.19 different images and eight correct base images as 2.61. Over all 32 images,
the mean cosine distance between two images tells the same story (0.21 for RL against 0.29 for the base model). Prompts on which RL gains more also lose more variety, but the link is weak (Spearman
correlation 0.24). So best-of-$N$ of the base model keeps more variety among its correct images, which matters
when a user wants several different images for one prompt.
\paragraph{How firm the first human check is.} \Cref{tab:judges} and \cref{fig:teaser}b of the paper come from this check, so here we ask
how much that result depends on its plan and on single labellers. Its plan (what is compared, the bootstrap and the only splits allowed) was
committed with a time stamp 15 minutes before the first label of labellers~2 and~3. They labelled 636 of the 796
images. The other 160 come from the 80 prompts that labeller~1 labelled the day before.
People judge the two images of a prompt differently for 96 prompts: 59 in favour of RL and 37
in favour of best-of-16. An exact sign test gives $p=0.032$, and so does a sign-flip test. Without the labels of
labeller~1, on the 318 prompts of labellers~2 and~3, the gap is $-6.6$ points ($-12.3$ to $-0.9$). Each labeller
alone finds RL ahead ($-1.3$, $-5.6$ and $-7.6$ points), but each alone has a wide interval. Leaving out labeller~2
or labeller~3 gives intervals that just include zero ($-5.5$, $-11.3$ to $+0.4$; $-4.2$, $-10.4$ to $+1.7$). The direction of the human
result is clear, and its size is uncertain.

\subsection{A second human check, fixed in advance}
\label[appendix]{sec:supp-humanr}
The first check was decided after the registered result, and 160 of its images had been labelled a day earlier with
the general rules only. So we repeated it on a new random draw, and we committed its plan with a time stamp before
the draw. For every test prompt, the plan draws one new random image of \grpo and 16 new random seeds of \bstar, from
the 32 stored images of each model. Among these 16 base images, the VLM picker and its 38B version each choose one
image. Because both checks draw from the same stored images, 140 of the 398 picks of the 8B picker and 4 of the RL
images are the same images as in the first check. They were labelled again. Only labellers~2 and~3 took part. Each
labelled all images of about half of the prompts (202 and 196). The two halves were balanced over the prompt sets,
the hard prompts and the position prompts. Each labeller also labelled 25 prompts of the other half, so that 50
prompts have two labels. Together they made 1,196 labels, and on the 133 images labelled twice they agree on 130
($\kappa=0.94$).

The planned test is $D$, the pick of the 8B picker minus the RL image, and the check succeeds if its 95\% interval
lies below zero. It does: $D=-6.5$ points ($-11.1$ to $-2.0$, $p=0.006$, \cref{tab:supp-humanr}). The rest agrees
with the first check, and this time the plan named the splits in advance: position prompts or not, the two prompt
sets, and hard prompts or not. The gap lies on the position prompts ($-39.2$ points, $-51.9$ to $-27.8$) and not on
the others ($+1.6$, $-2.8$ to $+6.0$). So the finding that RL is ahead mainly on positions, which the first check made
without a plan, holds in a planned split. The gap also lies on the hard prompts ($-14.7$ against $-1.2$). RL stays
ahead on the 140 new-object prompts, whose objects the GenEval reward cannot recognise ($-5.7$ points, $-10.7$ to
$-1.4$), as on the 258 known-object prompts ($-7.0$, $-13.6$ to $-0.4$). In both checks, all of the gap on new objects comes from
their 19 position prompts; in the first check, the gap on new objects was $-5.0$ points (\cref{tab:supp-types,tab:supp-humanr-types}). So on
new objects, too, the lead of RL is a lead on positions. Over all prompts of this check, the registered rule and the detector
judge report about $-10$ points, and the VLM judge puts best-of-16 ahead by 3.5 points. According to people, the 38B picker chooses better images than the 8B
picker ($+4.3$ points, $+1.0$ to $+7.5$). Its gap to one RL image, $-2.3$ points ($-7.0$ to $+2.5$), is no longer
significant, while the registered rule still reports $-5.8$ points on the same images. On the position prompts, RL
stays far ahead of the 38B picker too ($-34.2$ points, $-45.6$ to $-22.8$). On the other prompts, the 38B picks are
ahead of one RL image ($+5.6$ points, $+0.9$ to $+10.3$). Over all draws of the stored images, the registered rule
finds the 38B picker only $0.8$ points better than the 8B picker (see \emph{A stronger picker} above), so part of its
gain on this draw may be chance.

After finishing the labels, labeller~2 reported a possible mistake near one image. The seven images around it were shown to
labeller~2 again in a separate set, and one label changed. The planned analysis keeps the first labels. With the
changed label, $D=-6.3$ points ($-10.8$ to $-1.8$). \Cref{tab:supp-humanr-types} splits this check by prompt type.

\paragraph{The judges against people, in both checks.}
On the images of this check, the detector judge makes the gap 3.8 points larger than people see ($-1.0$ to $+8.5$),
and the VLM judge makes it 10.1 points smaller (5.8 to 14.3). So the two judges again err in opposite directions. The
error of the VLM judge is clear in both checks. The error of the detector judge has the same sign as in the first
check, where it was 5.3 points (0.8 to 9.8), but here it is within chance. Both checks cover the same 398 prompts, so
we can also average them, although the plan of the second check did not ask for this. On average, the detector judge makes the gap 4.5 points larger (0.6 to 8.4), and the VLM
judge makes it 8.0 points smaller (4.6 to 11.6). The reasons are the same as in the first check
(\cref{sec:analysis-judges}). In this check, the detector judge passes 75.6\% of the correct best-of-16 images but
86.7\% of the correct RL images, and the VLM judge passes 40.9\% of the best-of-16 images that people call wrong, but
only 6.7\% of the wrong RL images (\cref{tab:supp-judge2}). This is not a matter of its threshold. If we raise the threshold of the VLM judge from 0.5 to 0.82,
the value that best matches the 320 earlier labels of labeller~1, it still passes 40.0\% and 5.6\% of these wrong
images. A VLM judge of another family and one four times as large are lenient in the same way
(\cref{sec:supp-judge2}).

\begin{table}[t]
\centering\small
\setlength{\tabcolsep}{2.5pt}
\begin{tabular}{@{}lccccc@{}}
\toprule
Judge & RL & 8B & 38B & $D$, 8B & $D$, 38B \\
\midrule
People & 77.6 & 71.1 & 75.4 & \m6.5 \ci{\m11.1}{\m2.0} & \m2.3 \ci{\m7.0}{\p2.5} \\
Rule & 67.1 & 57.0 & 61.3 & \m10.1 \ci{\m14.6}{\m5.3} & \m5.8 \ci{\m10.6}{\m1.0} \\
Detector & 68.8 & 58.5 & 63.1 & \m10.3 \ci{\m15.1}{\m5.5} & \m5.8 \ci{\m10.8}{\m0.8} \\
VLM & 74.6 & 78.1 & 81.9 & \p3.5 \ci{\m0.3}{\p7.5} & \p7.3 \ci{\p3.5}{\p11.3} \\
\bottomrule
\end{tabular}
\caption{\textbf{The second human check}, on a new random draw for all 398 test prompts. RL: one random image of \grpo.
8B and 38B: the image that the VLM picker or its 38B version chooses among the same 16 random images of \bstar.
Shares of correct images in \%, and $D$ = picked image minus RL image in points (negative: RL is better), with 95\%
intervals. Rows: people, the registered rule (both judges must pass), and each automatic judge alone.}
\label{tab:supp-humanr}
\end{table}

\subsection{A human check of the perfect picker}
\label[appendix]{sec:supp-humano}
The exchange rate with a perfect picker rests on the automatic judges, because the perfect picker uses their verdicts.
So we asked people to judge its choices, and committed the plan with a time stamp before any of its tasks or labels
existed. For every test prompt we fixed a random order of the 32 base images. The perfect picker's choice among the
first $N$ images is the first image in this order that both judges pass, or the first image if none passes, as
in the main paper. So two images per prompt, the first image and the first one that passes, give its choice for every $N$ from 1 to
32. Labellers~2 and~3 judged these images and one new random RL image per prompt. They did not see the model, and
the images came in random order. Each prompt went to the labeller who had it in the second human check. On the 123 images that both labelled they agree on 122 ($\kappa=0.98$). Of the images in this check, 71 had been
labelled before, some of them more than once. Of the 153 pairs of a new and an earlier label of the same image, 150 agree.

\Cref{tab:supp-humano} gives the result. People find one RL image correct for 79.1\% of the prompts, and the perfect
picker's choice among 8 base images for 69.8\%. The planned test, the choice at $N=8$ minus the RL image, is
$D=-9.3$ points ($-14.1$ to $-4.5$, $p<0.001$), so by the test fixed in advance this picker needs more than 8 base images when people judge.
The curve rises slowly and is still below the RL image at $N=32$ ($-4.8$ points, $-9.5$ to $0.0$), and in 96\% of the
bootstrap rounds it does not reach it within 32 images. The planned split by position shows where this comes from.
On the 319 prompts without a position, the choice among 8 base images is as good as one RL image ($+0.3$ points,
$-4.4$ to $+5.0$), and among 32 it is level or slightly better ($+4.1$, $-0.6$ to $+9.1$). On the 79 position prompts it stays
far behind ($-48.1$ points at $N=8$ and $-40.5$ at $N=32$).

Why do people and the judges differ? Not mainly on the base images: by the judges the perfect picker's choice among 8
images is correct for 68.8\% of the prompts, and by people for 69.8\%. People fail 8.4\% of the choices that both
judges pass, but they pass some first images that the judges fail, and the two errors nearly cancel. The difference
lies on the RL side. The judges pass only 83.5\% of the RL images that people call correct, so people find RL correct
for 79.1\% of the prompts where the judges find 66.8\%. On the position prompts the gap is largest: people find RL
correct for 92.4\% of them and the judges for 73.4\%, while about one in five of the base images that the judges pass there is wrong. So, for this picker, the
exchange rate of about 8 holds by our judges. By people, it holds for the prompts without a position, and on
positions even 32 base images chosen this way fall far short of one RL image.

The plan also had a second part, in which a person who failed the first image of a prompt was shown the
next images of that prompt. On the day of this check the labellers had time for only 26 of these images. They
finished the second part six days later, and \cref{sec:supp-search} reports it. The results
above are from the first part alone. They measure the perfect picker of the paper, which uses the judges' verdicts.
The second part measures a picker that knows what people would say, and such a picker does better.

\begin{table*}[t]
\centering\small
\setlength{\tabcolsep}{8pt}
\begin{tabular}{@{}lccc@{}}
\toprule
 & All prompts (398) & Prompts without a position (319) & Position prompts (79) \\
\midrule
\multicolumn{4}{@{}l}{\emph{Perfect picker's choice among $N$ base images}} \\
$N=1$ & 58.0 & 65.8 & 26.6 \\
$N=2$ & 61.1 & 69.0 & 29.1 \\
$N=4$ & 65.8 & 73.0 & 36.7 \\
$N=8$ & 69.8 & 76.2 & 44.3 \\
$N=16$ & 72.9 & 79.3 & 46.8 \\
$N=32$ & 74.4 & 79.9 & 51.9 \\
\midrule
One new RL image & 79.1 & 75.9 & 92.4 \\
\midrule
$D$ at $N=8$ & \m9.3 \ci{\m14.1}{\m4.5} & \p0.3 \ci{\m4.4}{\p5.0} & \m48.1 \ci{\m59.5}{\m36.7} \\
$D$ at $N=32$ & \m4.8 \ci{\m9.5}{0.0} & \p4.1 \ci{\m0.6}{\p9.1} & \m40.5 \ci{\m51.9}{\m29.1} \\
\bottomrule
\end{tabular}
\caption{\textbf{The perfect picker judged by people.} Share of prompts (\%) on which people find the image correct.
The upper rows give the choice of the perfect picker among the first $N$ base images; this picker uses the verdicts
of both automatic judges. $D$: its choice minus one new RL image, in points
(negative: RL is better), with 95\% intervals.}
\label{tab:supp-humano}
\end{table*}

\subsection{Other checkpoints, searches and judges}
\label[appendix]{sec:supp-other}
The main comparison uses one base model, one RL checkpoint, plain best-of-$N$ and our own two judges. This subsection changes each
of these in turn.

\paragraph{Two more RL checkpoints.}
The plan lists two more public Flow-GRPO checkpoints of SD3.5-M as exploratory: a second run on the GenEval
reward by other authors (\path{shi-labs/SD3.5M-ind-expert-GenEval}), and a run without the KL penalty by the Flow-GRPO
authors (\path{jieliu/SD3.5M-FlowGRPO-GenEval-without-KL}). We drew seeds 0--15 for the 398 test prompts with the settings of
\grpo. We then compared one image of each checkpoint with the same best-of-16 of \bstar with the VLM picker
(\cref{tab:supp-ckpt}). \grpo also uses seeds 0--15 here, which gives 68.5\% instead of the 68.0\% of all 32 seeds.
The second run behaves like \grpo. One of its images is correct for 69.6\% of the prompts, the exchange rate with a
perfect picker is 8.7, and best-of-16 is 10.9 points behind it. On our prompts, the run without the KL penalty is less accurate. One of its
images is correct for 58.1\% of the prompts, and best-of-16 matches it ($+0.7$ points). It loses mostly on the
prompts without a position (54.2\% against 66.1\% for \grpo) and keeps most of its lead on positions (73.7\%
against 78.2\% for \grpo and 40.3\% for best-of-16). Without the penalty, a model is free to move away from what the
base model can draw (\cref{prop:tilt}), but this run did not become stronger where the base model is weak. The two
runs may also differ in other ways, such as the length of training, so this is an observation and not a test of the
penalty. For the three checkpoints, the detector judge alone gives the same picture ($D=-10.2$, $-12.3$ and $-3.5$
points), and the VLM judge alone again sees less of the gap ($+1.8$, $+0.7$ and $+11.5$ points).

\begin{table}[t]
\centering\small
\setlength{\tabcolsep}{3pt}
\begin{tabular}{@{}lcccc@{}}
\toprule
RL model & One & $\nstar$ & $D$ & Positions / other \\
\midrule
\grpo (main) & 68.5 & 7.9 & \m9.7 \ci{\m13.3}{\m6.2} & \m37.8 / \m2.8 \\
Second run & 69.6 & 8.7 & \m10.9 \ci{\m14.7}{\m7.1} & \m42.6 / \m3.0 \\
Without KL & 58.1 & 3.5 & \p0.7 \ci{\m3.6}{\p5.0} & \m33.4 / \p9.1 \\
\bottomrule
\end{tabular}
\caption{\textbf{Three Flow-GRPO checkpoints of SD3.5-M}, seeds 0--15, under the registered rule. One: the share of
prompts on which one image is correct, in \%. $\nstar$: the exchange rate with a perfect picker. $D$: best-of-16 of
\bstar with the VLM picker (58.8\%) minus one image of the RL model, in points (negative: RL is better), with its
95\% interval. Positions / other: $D$ on the 79 position prompts and on the 319 others.}
\label{tab:supp-ckpt}
\end{table}

\paragraph{Noise search.}
Best-of-16 draws 16 independent starting noises. A search can spend the same 16 images differently: it can start new images near the noise of the best image so far,
similar to the zero-order search of~\cite{ma2025inference}. This search is not in the registered plan. We added it to see whether a
better use of the same samples closes the gap, and we fixed its design before generating the first image. The search
starts from the noises of seeds 0--3. In each of three more rounds, it draws four new noises
$\sqrt{1-\lambda^2}\,z+\lambda\,\epsilon$, where $z$ is the noise of the image that the VLM picker likes best so far
and $\epsilon$ is fresh Gaussian noise. It returns the image with the highest picker score among all 16. With
$\lambda=1$ the search is plain best-of-16. The model, the sampler and the picker are those of best-of-16. We chose
$\lambda$ on the 87 dev prompts as the value below 1 with the most correct outputs under the registered rule. The
shares were 56.3\%, 59.8\%, 58.6\% and 54.0\% for $\lambda=0.25$, 0.5, 0.75 and 1, so we used $\lambda=0.5$ on the
test prompts. One dev prompt is worth 1.15 points, so this choice is not firm. On the 398 test prompts the search is
correct for 60.6\%, which is 1.8 points more than best-of-16 with the same picker ($-1.7$ to $+5.3$) and 7.5 points
less than one RL image (3.4 to 11.4). Under the VLM judge alone it gains nothing ($-0.8$ points). On the
position prompts it reaches 48.1\%, against 40.3\% for best-of-16 and 78.0\% for RL. The returned image comes from
rounds 0, 1, 2 and 3 for 76, 104, 102 and 116 prompts. As with the larger pickers, the limit is the picker:
searching harder for the images it likes does not find many more correct ones. By chance, this search and the search
over previews above are both correct for exactly 241 prompts, although they disagree on 76. As a check of the code,
we compared 192 images of round 0 on dev and 192 on test with the stored images of \bstar for the same seeds, and all
were identical, file for file.

\ifflux
\paragraph{FLUX.1-dev.}
The FLUX comparison uses \fluxPromptText{} with \fluxSeedText, the same judges and the same VLM picker. The
prompt set and the number of seeds follow a rule fixed before the run: all 398 prompts if their images were finished
by a date fixed in advance, otherwise the 200 registered ones, and 32 base seeds where they were finished. All these
images were finished by that date. One image is correct for \fluxBaseOne\% of the prompts with FLUX.1-dev and for \fluxRLOne\%
with its RL adapter. The exchange rate with a perfect picker is \fluxNstarPerfect. Best-of-16 with the VLM picker
reaches \fluxBonSel\%. Its difference from one RL image is \fluxD{} points under the registered rule, \fluxDdet{}
under the detector judge and \fluxDvlm{} under the VLM judge. The exchange rate above uses only
$N = 1, 2, 4, \dots, 32$. If we evaluate the curve at every $N$ from 1 to 32, as for SD3.5-M, it is 6.3. Best-of-32 reaches 63.9\%,
still below one RL image. On the 200 registered prompts with 16 base seeds, which is the sample that the plan names
first, the difference is $-11.1$ points ($-16.9$ to $-5.5$). The picture is the same as for SD3.5-M. Most of the gap
comes from position prompts ($-29.1$ points, against $-2.9$ on the other types). As for SD3.5-M, the VLM judge alone
sees no gap (\fluxDvlm{} points here, against $+2.0$ for SD3.5-M), and the human checks show that this judge accepts many wrong picks of the VLM picker
(\cref{sec:analysis-judges}). We have no
human labels for FLUX, so its gap rests on the automatic judges alone. A human check of FLUX.1-dev is planned but
not yet labelled (\cref{sec:supp-rules}).
\fi

\paragraph{Other detector judges.}
Is the bias of our detector judge shared by other detector-based judges? On the known-object prompts, every image
also has the verdict of the official GenEval code and of the GenEval reward,
computed as in the RL code (\cref{tab:supp-gate0}). We compute the reward twice: once on the image itself and once
after the JPEG compression that the reward client applies during RL training. \Cref{tab:supp-judges} compares these
judges with people on the images of both human checks. All of them are detectors followed by rules, like our detector judge, and all show its uneven
sensitivity. The official code and the reward used in training pass 56--70\% of the correct best-of-16 images but
81--86\% of the correct RL images, and they report gaps of 10 to 17 points where people see 6 to 7. The reward is the most uneven of all. The VLM
judge again sees no gap, or puts best-of-16 ahead. So the bias of
\cref{sec:analysis-judges} is not a property of our own detector judge alone: the official GenEval code and the reward
show it too, and the reward most of all. The official code and the reward, however, use the same detector and the
same colour classifier, so they do not show that a judge with no part in common with the reward would be biased.

\paragraph{Where the detector judge is uneven.}
To see where this bias comes from, we split the prompts of both human checks into the 167 whose type asks for a
colour (three colour bindings, colour and position, colours, and colour attribution) and the 231 others. We did not
plan this split. On the prompts with colours, our detector judge passes 81.8\% of the correct RL images but only
60.7\% of the correct best-of-16 images in the first check, a difference of 21.1 points (11.6 to 30.6), and 76.2\%
against 54.6\% in the second check (21.6 points, 11.9 to 31.2). On the other prompts the difference is only 4.8 points
(0.5 to 9.4) and 5.4 points (0.8 to 10.4). The prompts with three objects and no colour also show crowded scenes, yet
there the judge passes 94.1\% of the correct images of both models in the first check. For colours, every
detector-based judge in \cref{tab:supp-judges} uses the same CLIP classifier as the GenEval reward, which RL was
trained to satisfy. This shared classifier is not the only cause, however. Our detector judge with a colour
classifier trained elsewhere is still uneven on the prompts with colours, while VLM judges are much less uneven
there or not clearly so (\cref{sec:supp-judge2}). So most of the bias comes from the kind of colour check that these judges share with the reward,
a check that names the colour of each detected object. A detector that took no part in training does not alone make
a judge neutral. The
intervals are 95\% intervals from 10,000 bootstrap rounds over prompts. \Cref{tab:supp-misses} in
\cref{sec:supp-stats} splits the failures of our detector judge by the check that fails.

\begin{table*}[t]
\centering\small
\setlength{\tabcolsep}{4pt}
\begin{tabular}{@{}lccccccc@{}}
\toprule
 & \multicolumn{3}{c}{First human check} & & \multicolumn{3}{c}{Second human check} \\
\cmidrule(lr){2-4}\cmidrule(l){6-8}
Judged by & $D$ (points) & Correct & Wrong & & $D$ (points) & Correct & Wrong \\
\midrule
People & \m5.8 \ci{\m12.8}{\p1.2} & -- & -- & & \m7.0 \ci{\m13.6}{\m0.4} & -- & -- \\
Both judges (registered rule) & \m10.5 \ci{\m17.1}{\m3.9} & 68.4 / 81.4 & 10.4 / \phantom{0}6.6 & & \m9.3 \ci{\m15.9}{\m2.7} & 64.9 / 77.3 & 12.5 / \phantom{0}4.7 \\
Detector judge & \m11.2 \ci{\m17.8}{\m4.3} & 70.4 / 85.0 & 13.2 / \phantom{0}8.8 & & \m9.3 \ci{\m16.3}{\m2.7} & 66.2 / 79.7 & 16.3 / \phantom{0}7.0 \\
Official GenEval & \m10.9 \ci{\m17.4}{\m4.3} & 70.4 / 83.8 & 15.1 / 12.1 & & \m10.1 \ci{\m16.7}{\m3.5} & 68.2 / 84.9 & 20.2 / \phantom{0}7.0 \\
GenEval reward, RL code & \m16.7 \ci{\m23.3}{\m10.1} & 61.8 / 85.6 & 13.2 / \phantom{0}8.8 & & \m14.7 \ci{\m21.3}{\m8.1} & 57.8 / 83.1 & 18.3 / \phantom{0}3.5 \\
\quad after JPEG, as in training & \m15.9 \ci{\m22.1}{\m9.3} & 58.6 / 81.4 & 14.2 / \phantom{0}9.9 & & \m17.1 \ci{\m23.6}{\m10.1} & 56.5 / 83.1 & 15.4 / \phantom{0}4.7 \\
VLM judge & \p1.9 \ci{\m3.9}{\p7.8} & 92.8 / 92.2 & 31.1 / 16.5 & & \p7.0 \ci{\p1.2}{\p12.8} & 90.3 / 90.7 & 39.4 / \phantom{0}7.0 \\
\bottomrule
\end{tabular}
\caption{\textbf{Automatic judges against people on the 258 known-object prompts of the two human checks.} $D$:
best-of-16 with the VLM picker minus one RL image, in points (negative: RL is better), with 95\% intervals. Correct:
of the images that people label correct, the share (\%) that the judge passes, given as best-of-16 / RL. Wrong: the
same for the images that people label wrong. A fair judge gives two equal numbers in each pair. Every detector-based judge passes far fewer of the correct best-of-16 images
than of the correct RL images.}
\label{tab:supp-judges}
\end{table*}

\begin{table*}[t]
\centering\small
\setlength{\tabcolsep}{6pt}
\begin{tabular}{@{}lcccccc@{}}
\toprule
 & & & \multicolumn{2}{c}{Perfect picker} & & \\
\cmidrule(lr){4-5}
Prompt type & Prompts & One base & Best of 16 & Best of 32 & One RL & $\nstar$ \\
\midrule
\multicolumn{7}{@{}l}{\emph{Known objects}} \\
Count of five & 18 & 12.0 & 64.5 & 72.2 & 54.0 & 8.5 \ci{4.7}{$>$32} \\
Two counted classes & 60 & 9.9 & 53.2 & 61.7 & 16.1 & 1.7 \ci{1.3}{2.4} \\
Three objects & 60 & 74.3 & 97.8 & 98.3 & 82.9 & 1.7 \ci{1.1}{2.7} \\
Three colour bindings & 60 & 15.1 & 66.1 & 75.0 & 46.0 & 5.5 \ci{4.0}{7.5} \\
Colour and position & 60 & 14.2 & 51.6 & 58.3 & 73.4 & $>$32 \\
\multicolumn{7}{@{}l}{\emph{New objects}} \\
One object & 23 & 99.5 & 100.0 & 100.0 & 100.0 & -- \\
Two objects & 25 & 95.0 & 100.0 & 100.0 & 99.0 & 1.9 \ci{1.5}{8.0} \\
Counting & 26 & 48.1 & 92.1 & 96.2 & 91.8 & 15.3 \ci{5.8}{$>$32} \\
Colours & 26 & 86.9 & 96.2 & 96.2 & 95.9 & 3.7 \ci{1.8}{$>$32} \\
Position & 19 & 22.0 & 78.4 & 84.2 & 92.4 & $>$32 \ci{10.4}{$>$32} \\
Colour attribution & 21 & 37.9 & 77.1 & 85.7 & 75.3 & 13.5 \ci{5.8}{24.8} \\
\midrule
All but positions & 319 & 47.5 & 79.9 & 84.6 & 65.5 & 3.9 \ci{3.2}{4.9} \\
The two position types & 79 & 16.1 & 58.1 & 64.6 & 78.0 & $>$32 \\
\bottomrule
\end{tabular}
\caption{\textbf{The exchange rate with a perfect picker, by prompt type.} Share of prompts with a correct image (\%)
under the registered rule: one base image, best-of-16 and best-of-32 of the base model with a perfect picker, and one
RL image. $\nstar$: base images per prompt needed to match one RL image, with its 95\% interval (2,000 bootstrap
rounds). $>$32: not reached within 32 images. No interval is given when even the lower end of the interval is above 32. --: both models are correct for almost every prompt. The split by type was not planned.}
\label{tab:supp-nstar-types}
\end{table*}

\begin{table}[t]
\centering\small
\setlength{\tabcolsep}{3pt}
\begin{tabular}{@{}L{0.42\linewidth}ccc@{}}
\toprule
Base image (398 prompts) & Base & RL & Base minus RL \\
\midrule
\multicolumn{4}{@{}l}{\emph{SD3.5-M}} \\
Best-of-16 pick, first check & 5.45 & 5.44 & \p0.01 \ci{\m0.02}{\p0.05} \\
Best-of-16 pick, second check & 5.41 & 5.45 & \m0.03 \ci{\m0.07}{\p0.00} \\
\quad with the 38B picker & 5.42 & 5.45 & \m0.03 \ci{\m0.06}{\p0.01} \\
One random base image & 5.45 & 5.44 & \p0.01 \ci{\m0.02}{\p0.05} \\
\multicolumn{4}{@{}l}{\emph{FLUX.1-dev}} \\
Best-of-16 pick & 5.68 & 5.74 & \m0.06 \ci{\m0.08}{\m0.03} \\
One random base image & 5.68 & 5.74 & \m0.06 \ci{\m0.08}{\m0.03} \\
\bottomrule
\end{tabular}
\caption{\textbf{Image quality by the LAION aesthetic score} (1 to 10, higher is better), one image of each kind per
prompt. Base: the image of the base model named in the row; RL: one random RL image of the same prompt. The last
column gives the paired difference with its 95\% interval. Rows with the same numbers differ only in the third
decimal.}
\label{tab:supp-quality}
\end{table}

\section{What a KL penalty predicts}
\label[appendix]{sec:supp-tilt}

\grpo was trained with a KL penalty that keeps it close to the base model. Here we state what such a penalty implies
if training reached its optimum. The result is known in other settings~\cite{uehara2024entropy,korbak2022rlkl}. We
restate it because it gives the prediction that \cref{sec:analysis-mech} tests.

Consider a single prompt. Let $p_0$ be the distribution of images that the base model draws for it, and let
$r(x)\in\{0,1\}$ say whether image $x$ is correct. RL with a KL penalty of weight $\beta$ maximises
\begin{equation}
  J(p) \;=\; \mathbb{E}_{x\sim p}[r(x)] \;-\; \beta\,\KL(p\,\|\,p_0).
\end{equation}

\begin{proposition}
\label{prop:tilt}
The maximum is reached by the tilted distribution $p^\star(x) \propto p_0(x)\,e^{r(x)/\beta}$. Two consequences follow.
(i) $p^\star$ puts no mass where $p_0$ puts none, so it cannot produce an image that the base model could never
produce.
(ii) If $q$ is the pass rate of the base model on the prompt, the pass rate $p$ at the optimum satisfies
$\logit p = \logit q + 1/\beta$. The shift $1/\beta$ is the same for every prompt.
\end{proposition}

\begin{proof}
We can write $J(p) = -\beta\,\KL\!\left(p \,\|\, p^\star\right) + \beta \log Z$ with $Z=\mathbb{E}_{p_0}[e^{r/\beta}]$,
which is largest when $p=p^\star$. Part (i) follows from the form of $p^\star$. For part (ii), $p^\star$ gives the
correct images a total mass of $q e^{1/\beta}/Z$ and the wrong images a mass of $(1-q)/Z$, so the odds of a correct
image are \mbox{$e^{1/\beta}\,q/(1-q)$}.
\end{proof}\par Part (i) says little for image models, which give every image some mass. A more useful form follows from $Z\ge 1$: $p^\star(x)\le e^{1/\beta}\,p_0(x)$ for every image $x$. So a tilt can make no image, and no pass rate, more than $e^{1/\beta}$ times as large as under the base model.

\paragraph{The exchange rate under a tilt.}
Let $c$ be the common shift in log-odds, the same for every prompt. A prompt with base pass rate $q$ then has the RL pass rate $p = e^c q / (1 - q + e^c q)$.
With independent draws, the pass@$N$ of the base model is $1-(1-q)^N$, which is close to $Nq$ when $Nq$ is small. For
small $q$, $p$ is close to $e^c q$, so the two meet at $N\approx e^c$. For larger $q$ the crossing comes earlier, and
an average over prompts mixes both cases. Best-of-$N$ also only reweights the base distribution, and its KL divergence
from the base model is at most $\log N - (N-1)/N$~\cite{beirami2025bon}. In this idealised picture both options stay
within what the base model can draw. They differ in how strongly they reweight it and in when the cost is paid.

\paragraph{How we test it.}
Real training differs from the proposition in three ways. The KL term is applied at each denoising step and not to
the distribution of final images. The training reward is a soft GenEval score and not a 0/1 verdict. And training
stops before the optimum. So we use the proposition only as a point of reference, and not to estimate $\beta$.
We estimate a single shift $c$ as the mean of the per-prompt changes in log-odds, each weighted by the inverse of its
variance. To avoid infinite log-odds, we add 0.5 to each count. We measure how much the prompts disagree with this shift by the $I^2$
statistic of meta-analysis~\cite{higgins2002i2}. Finally, we regress the shift on the base log-odds. For this we split
the seeds into two halves, so that the same counts do not appear on both axes. We compare the slope with its
distribution under a simulated exact tilt.

\paragraph{Result.}
\Cref{tab:supp-tilt} gives the estimates. The common shift is $c=1.80$, so $e^c=6.1$. The prompts disagree strongly
with a common shift ($I^2=0.69$, against 0 under an exact tilt), and the shift is larger for prompts on which the base
model is weak (slope $-0.23$, against $-0.06$ under an exact tilt). A common shift predicts an accuracy of 61.5\% for
the RL model, where 68.0\% is observed, and an exchange rate of 4.5, where 7.2 is observed against the same base
model. Part of this shortfall comes from the way we estimate $c$. When we simulate the RL counts from an exact tilt and
estimate $c$ in the same way, the prediction falls short by only 1.7 points (1.3 to 2.1 in 95\% of 1,000
simulations), not by 6.5 points as here. So on the prompts that are hard for the base model, the RL checkpoint gains more than a uniform reweighting
would give.

\begin{table}[t]
\centering\small
\setlength{\tabcolsep}{3pt}
\begin{tabular}{@{}lccc@{}}
\toprule
Prompts & Shift $c$ & $I^2$ & Slope \\
\midrule
All & 1.80 \ci{1.60}{1.99} & 0.69 \ci{0.64}{0.74} & \m0.23 \ci{\m0.30}{\m0.16} \\
Known & 1.67 \ci{1.43}{1.89} & 0.74 \ci{0.68}{0.78} & \m0.22 \ci{\m0.33}{\m0.11} \\
New & 2.28 \ci{1.94}{2.61} & 0.52 \ci{0.38}{0.60} & \m0.49 \ci{\m0.62}{\m0.36} \\
\bottomrule
\end{tabular}
\caption{\textbf{How the log-odds of a correct image change from \base to \grpo.} Shift $c$: the common shift in
log-odds. $I^2$: the share of the variation between prompts that sampling noise does not explain (0 under an exact
tilt). Slope: the slope of the shift against the log-odds of the base model. Known, New: known-object and new-object
prompts. Small numbers are 95\% intervals.}
\label{tab:supp-tilt}
\end{table}

\paragraph{A second point of reference.}
The proposition describes the optimum of the KL-penalised objective, which need not be where GRPO converges.
Flow-GRPO normalises the rewards of each group by their mean and standard deviation. For exact GRPO with a pass/fail
reward and a KL penalty, the pass rate converges to a fixed point that solves
$\logit p = \logit q + a/\sqrt{p(1-p)+\epsilon}$ (\cite{mroueh2025grpo}, Theorem~2). We fitted $a$ by maximum
likelihood on the per-prompt counts. This model fits our data worse than the common shift (log-likelihood $-7700$
against $-5826$ for a single common shift fitted in the same way) and predicts an accuracy of 49.9\% for the RL model. Fitted in this way, the common shift is 2.35 instead of 1.80, and it matches the observed 68.0\% by construction.

\section{More related work}
\label[appendix]{sec:supp-related}

This section extends \cref{sec:related} with work on five topics that bear on our findings.

\paragraph{Spatial relations and layout.}
Spatial relations are a known weak point of text-to-image models. VISOR~\cite{gokhale2022visor} measures them with a
detector and finds that models often draw both objects but in the wrong place, and SPRIGHT~\cite{chatterjee2024spright}
traces part of the problem to training captions that rarely describe positions. T2I-CompBench++~\cite{huang2025t2icompbench},
DPG-Bench~\cite{hu2024ella} and GenAI-Bench~\cite{li2024genaibench} also test relations, with more varied wording
than GenEval. Training-free methods steer attention towards each object~\cite{chefer2023attend} or towards boxes that
a user or a language model provides~\cite{chen2024layoutguidance,xie2023boxdiff,feng2023layoutgpt,lian2024lmd,phung2024refocusing},
and a learned loss can steer towards a requested relation at test time~\cite{yiflach2026learntosteer}. A reward model
for spatial relations can also drive RL~\cite{tang2026spatialscore}. Its authors report that Flow-GRPO trained on
GenEval generalises poorly to complex relations, and GenEval-trained Flow-GRPO also loses accuracy when its prompts
are reworded~\cite{wang2026promptrl}. Our finding that RL is ahead mostly on position prompts fits this work: the
base model rarely draws a correct layout, so a picker has less to find. Our prompts follow the GenEval templates
used to train RL, so they do not test complex or reworded relations.

\paragraph{The starting noise.}
The starting noise fixes much of the layout. Some seeds place objects in typical regions of the
image~\cite{xu2025goodseed,li2025seeds}, and choosing or optimising the noise improves compositional and spatial
accuracy~\cite{guo2024initno,qi2024noises,li2025seeds}. Noise Hypernetworks~\cite{eyring2025hypernoise} train a
network to move the noise where a reward-guided optimisation would, and so turn test-time compute into training
compute. Our finding that RL secures half of its gain within the first 3.5 of 40 steps agrees with this picture.

\paragraph{Reward-guided sampling.}
Twisted sequential Monte Carlo~\cite{wu2023tds} and value-based decoding without gradients~\cite{li2025svdd} also
spend compute at inference to sample from a distribution tilted towards a reward, and two tutorials review these
methods and their relation to RL fine-tuning~\cite{uehara2024rltutorial,uehara2025tutorial}. When the samples of one
prompt repeat the same failure, best-of-$N$ with a fixed prompt stops improving~\cite{kim2026pris}.

\paragraph{Judges and pickers.}
LLM judges recognise and favour their own outputs~\cite{panickssery2024self}, and multimodal judges favour their own
outputs and those of models of the same family~\cite{koyama2026philautia}. Our VLM picker and VLM judge both use a Qwen3
language model, which may make the judge pass more wrong picks. Concurrent work on counting finds that
every automatic verifier it tests shrinks the difference that people see between two generators, and that some
reverse it~\cite{li2026verifierchoice}; we find the same for a VLM judge and the opposite for detector judges. Reward
models that serve both as best-of-$N$ pickers and as RL rewards are weak at relations and
counting~\cite{su2026visualreasoning}. Human evaluations of text-to-image models should state their protocol so that
others can check and repeat it~\cite{otani2023verifiable}; \cref{sec:supp-prompts} gives ours.

\paragraph{Mixing models across steps.}
Swapping a base model and an RL-tuned model across sampling steps suggests that the early steps carry most of the
change~\cite{barcelo2024hrf}, although RL training has been found less effective at the earliest and at the latest
steps~\cite{yan2026adascope}.
Running the base model early and a tuned or distilled model late keeps more of the base model's
diversity~\cite{sorokin2025imagerefl,gandikota2025diversity}. Trajectory stitching~\cite{pan2025tstitch}, experts for
different noise levels~\cite{balaji2022ediffi} and guidance applied over a limited
interval~\cite{kynkaanniemi2024interval} rest on similar ideas. Our experiments that switch models during sampling
add guidance as a reference.

\section{Statistical details}
\label[appendix]{sec:supp-stats}

\begin{table*}[t]
\centering\small
\setlength{\tabcolsep}{5pt}
\begin{tabular}{@{}lc ccc cc c@{}}
\toprule
 & & \multicolumn{3}{c}{Correct (\%)} & \multicolumn{2}{c}{$D$ (points)} & First check \\
\cmidrule(lr){3-5}\cmidrule(lr){6-7}
Prompt type & Prompts & One RL & 8B pick & 38B pick & 8B picker & 38B picker & $D$ (points) \\
\midrule
\multicolumn{8}{@{}l}{\emph{Known objects}} \\
Count of five & 18 & 61.1 & 55.6 & 72.2 & \m5.6 \ci{\m33.3}{\p22.2} & \p11.1 \ci{\m16.7}{\p38.9} & 0.0 \\
Two counted classes & 60 & 15.0 & 33.3 & 48.3 & \p18.3 \ci{\p6.7}{\p30.0} & \p33.3 \ci{\p18.3}{\p46.7} & \p18.3 \\
Three objects & 60 & 85.0 & 85.0 & 85.0 & 0.0 \ci{\m11.7}{\p11.7} & 0.0 \ci{\m11.7}{\p11.7} & 0.0 \\
Three colour bindings & 60 & 80.0 & 71.7 & 75.0 & \m8.3 \ci{\m20.0}{\p5.0} & \m5.0 \ci{\m16.7}{\p5.0} & \m6.7 \\
Colour and position & 60 & 88.3 & 50.0 & 53.3 & \m38.3 \ci{\m51.7}{\m23.3} & \m35.0 \ci{\m48.3}{\m21.7} & \m36.7 \\
\midrule
\multicolumn{8}{@{}l}{\emph{New objects}} \\
One object & 23 & 100.0 & 100.0 & 100.0 & 0.0 \ci{0.0}{0.0} & 0.0 \ci{0.0}{0.0} & \p4.3 \\
Two objects & 25 & 100.0 & 100.0 & 100.0 & 0.0 \ci{0.0}{0.0} & 0.0 \ci{0.0}{0.0} & 0.0 \\
Counting & 26 & 96.2 & 96.2 & 96.2 & 0.0 \ci{\m11.5}{\p11.5} & 0.0 \ci{\m11.5}{\p11.5} & \m3.8 \\
Colours & 26 & 100.0 & 100.0 & 100.0 & 0.0 \ci{0.0}{0.0} & 0.0 \ci{0.0}{0.0} & 0.0 \\
Position & 19 & 94.7 & 52.6 & 63.2 & \m42.1 \ci{\m63.2}{\m21.1} & \m31.6 \ci{\m57.9}{\m5.3} & \m36.8 \\
Colour attribution & 21 & 95.2 & 95.2 & 90.5 & 0.0 \ci{\m14.3}{\p14.3} & \m4.8 \ci{\m14.3}{0.0} & 0.0 \\
\midrule
All prompts & 398 & 77.6 & 71.1 & 75.4 & \m6.5 \ci{\m11.1}{\m2.0} & \m2.3 \ci{\m7.0}{\p2.5} & \m5.5 \\
\bottomrule
\end{tabular}
\caption{\textbf{The second human check by prompt type.} Share of prompts (\%) on which people find the image correct:
one random RL image, and the image that the 8B or the 38B picker chooses among the same 16 base images. $D$: picked
image minus RL image, in points (negative: RL is better), with 95\% intervals. Last column: $D$ in the first human
check, from \cref{tab:supp-types}. On every type the two checks differ by less than 6 points. The split by type was
not planned.}
\label{tab:supp-humanr-types}
\end{table*}

This section adds four details to the sections above. We first say how the intervals were computed. We
then split the second human check by prompt type, and the failures of the detector judge by the check that fails.
Last, we ask which results remain after a correction for many tests. All four use only stored labels and verdicts,
and the last three were not planned.

\paragraph{How the intervals were computed.}
Every interval in the paper and in this supplement is a 95\% percentile interval from a bootstrap over prompts,
unless the text says otherwise. Each analysis starts its own random number generator (the default generator of
NumPy) with seed 0, and within one analysis all models and pickers use the same resampled prompts. The number of
rounds differs between analyses. The registered test and its variants by prompt set use 10,000 rounds. So do the
three human checks, the gap by prompt type (\cref{tab:supp-types}), the comparison of judges
(\cref{tab:supp-judges}), the larger pickers, the other RL checkpoints, the noise search, the variety and the quality
of the images, and FLUX.1-dev. The exchange rates and best-of-$N$ curves of SD3.5-M use 2,000 rounds
(\cref{sec:results-rate,tab:supp-ablations,tab:supp-nstar-types}), and so do the exchange rate judged by people, the tilt
analysis (\cref{tab:supp-tilt}) and the check of resolution. Under a single judge (\cref{tab:supp-ablations}),
$\nstar$ uses 1,000 rounds and $D$ uses 5,000, and the comparison of seeds in \cref{sec:analysis-mech} uses 1,000.
Because $\nstar$ can be infinite in a round, its interval takes the nearest round instead of interpolating between
two rounds. A $p$-value is twice the smaller share of rounds that fall on either side of zero, so a test in which no
round crosses zero is reported as $p<0.001$. More rounds change the intervals only a little. With 10,000 rounds, for
example, the interval of the main exchange rate is 5.8 to 10.2 instead of 5.9 to 10.0. A few results do not use the
bootstrap. The reach tests are exact McNemar tests, the test of deep sampling is an exact binomial test, the first
human check also has an exact sign test, and the share of the 20 prompts that the base model solves in 1,024 tries
has an exact (Clopper--Pearson) interval.

\paragraph{The second human check by prompt type.}
The plan of the second human check allowed only three splits: position prompts, prompt sets and hard prompts. A split
by prompt type is therefore exploratory, but it shows whether the pattern of the first check (\cref{tab:supp-types})
comes back on a new draw. \Cref{tab:supp-humanr-types} gives it, and the pattern does come back. With the 8B picker,
RL is far ahead on the two position types ($-38.3$ and $-42.1$ points), and best-of-16 is ahead on two counted
classes ($+18.3$ points, $+6.7$ to $+30.0$), by the same amount as in the first check. On three colour bindings
people again see only a small gap ($-8.3$ points, $-20.0$ to $+5.0$). On the seven other types the gap is zero, except for $-5.6$
points on the 18 prompts with a count of five. On every type, the gaps of the two checks differ by less than 6 points. The two draws are not fully
independent, because 140 of the 398 picks of the 8B picker are the same images. As in the first check, the
registered rule reports larger gaps than people on the two types with colour bindings: $-18.3$ and $-23.8$ points on
the same images, where people see $-8.3$ and 0.0. The 38B picker gives the same picture. More than half of its net gain
over the 8B picker (9 of 17 prompts) comes from two counted classes, where its lead over one RL image grows to
$+33.3$ points ($+18.3$ to $+46.7$). On the position types RL stays far ahead of it ($-35.0$ and $-31.6$ points).

\begin{table}[t]
\centering\small
\setlength{\tabcolsep}{4pt}
\begin{tabular}{@{}lcccc@{}}
\toprule
 & \multicolumn{2}{c}{First check} & \multicolumn{2}{c}{Second check} \\
\cmidrule(lr){2-3}\cmidrule(l){4-5}
Check that fails & Best-of-16 & RL & Best-of-16 & RL \\
\midrule
Object missing & 3.2 & 2.6 & 3.5 & 3.6 \\
Count & 3.2 & 0.7 & 4.2 & 1.3 \\
Colour & 14.3 & 6.6 & 15.9 & 7.8 \\
Position & 0.4 & 0.3 & 0.7 & 0.6 \\
\midrule
Any check & 21.1 & 10.3 & 24.4 & 13.3 \\
Correct images & 280 & 302 & 283 & 309 \\
\bottomrule
\end{tabular}
\caption{\textbf{Why the detector judge fails correct images.} Of the images that people label correct, the share
(\%) that the detector judge fails, split by the check that fails. Best-of-16: the image that the 8B VLM picker
chooses. An image that fails two checks (two images in each human check) is counted once, under the first of them in
the order of the rows. The colour check causes most of the difference between best-of-16 and RL. The split was not
planned.}
\label{tab:supp-misses}
\end{table}

\paragraph{Which check the detector judge fails.}
In \cref{sec:supp-other} we saw that the detector judge passes fewer of the correct best-of-16 images than of the
correct RL images, and that most of this difference lies on prompts with colours. The judge stores why it fails an
image, so we can also ask which of its checks fails: an object is missing, a count is wrong, the colour of an object
is wrong, or a position is wrong. \Cref{tab:supp-misses} splits the failures on the images that people label
correct. In both human checks, the colour check causes most of the difference. In the first check it fails 14.3\% of
the correct best-of-16 images and 6.6\% of the correct RL images. This difference of 7.7 points (3.5 to 12.0) is
71\% of the whole difference of 10.8 points. In the second check the colour check fails 15.9\% and 7.8\%, a
difference of 8.1 points (3.9 to 12.6) out of 11.1. On the prompts with colours, it accounts for almost all of the
difference (20.2 of 21.1 points in the first check, and 21.5 of 21.6 in the second). Missing objects and positions
add almost nothing. The rest, about a quarter, comes from counts: the judge miscounts 9 correct best-of-16 images and
2 correct RL images in the first check, and 12 and 4 in the second (differences of 2.6 points, 0.8 to 4.5, and 2.9
points, 0.8 to 5.3). The counts rest on the OWLv2 boxes, which took no part in training, so this smaller part does
not come from anything that the judge shares with the reward. The colour check, however, uses the CLIP classifier that the detector
judge shares with the GenEval reward (\cref{sec:supp-prompts}). So this split agrees with \cref{sec:supp-other}:
most of the uneven sensitivity of the detector judge sits in its colour check. The check in \cref{sec:supp-judge2}
shows that much of this unevenness remains when the colour check uses a classifier that shares nothing with the reward.

\paragraph{Many tests.}
The plan fixed one main test, the registered test of $D$. It also named two secondary tests, the McNemar test of reach
and a binomial test of deep sampling, each at $\alpha=0.05$ without correction. The second human check and the check
of the perfect picker each fixed one test in advance as well. All other results are exploratory. Many of them come
with a $p$-value or with an interval that excludes zero, and with so many tests some of them may be due to chance. To
see which results remain, we use the Holm correction, which keeps the chance of even one false finding in a group of
tests below 5\%. We group the tests by the question they answer and correct within each group
(\cref{tab:supp-holm}). For the first group we use 400,000 bootstrap rounds, because its $p$-values lie close to
0.05.

Most results remain. RL stays ahead on the position prompts in all three human checks and on the hard prompts in the
first two. The 38B picker stays better than the 8B picker (0.04), and the reach result and the uneven sensitivity of
the detector judge remain as well. Seven results do not remain. In the first human check, the gap seen by people
($p=0.03$) and the difference between the detector judge and people ($p=0.03$) rise to 0.08. Only the difference
between the VLM judge and people remains (0.03). This is why the second human check matters. It asks the same
question with a plan fixed in advance, so it needs no correction, and it gives $p=0.006$. In the split of the first
check by prompt type, the lead of best-of-16 on two counted classes and the lead of RL on positions with new objects
(both $p=0.01$) rise to 0.12. The second check finds both again, and there both remain after the same correction (0.03
and 0.002). Under the registered rule, the gaps on the count of five and on three objects rise from 0.02 and 0.03 to
0.12. With the 38B picker, the gap on positions with new objects rises from 0.03 to 0.23.

Three tests that were fixed in advance are not in \cref{tab:supp-holm}: the registered test of $D$ and the planned
tests of the second human check and of the perfect picker. They stay below 0.05 even if we add them to the group of
the first human check. The registered test and the check of the perfect picker then give 0.001, and the second human
check gives 0.03. Even if we put all
75 tests of \cref{tab:supp-holm} into a single group, 18 of them still remain, among them the gaps on the position
prompts in all three human checks and the reach result.

\begin{table}[t]
\centering\small
\setlength{\tabcolsep}{3pt}
\begin{tabular}{@{}lccc@{}}
\toprule
 & & \multicolumn{2}{c}{$p<0.05$} \\
\cmidrule(l){3-4}
Group of tests & Tests & before & after \\
\midrule
First human check, people and judges & 4 & 3 & 1 \\
Hard and position prompts, three checks & 12 & 6 & 6 \\
Prompt types, first check, people & 11 & 3 & 1 \\
Prompt types, registered rule & 11 & 8 & 6 \\
Prompt types, second check, 8B picker & 11 & 3 & 3 \\
Prompt types, second check, 38B picker & 11 & 3 & 2 \\
The 38B picker, second check & 4 & 3 & 3 \\
Perfect picker at $N=32$ & 3 & 1 & 1 \\
Reach & 4 & 3 & 3 \\
Uneven detector judge & 4 & 4 & 4 \\
\midrule
All of the above as one group & 75 & 37 & 18 \\
\bottomrule
\end{tabular}
\caption{\textbf{Results that remain after a correction for many tests.} Each row is a group of tests that answer one
question. Tests: the number of tests in the group. $p<0.05$: how many of them have $p<0.05$ before and after the Holm
correction within the group. The last row puts all 75 tests into one group. The $p$-values come from the bootstrap
with which each result was computed (400,000 rounds for the first group), or from an exact McNemar test for reach.
The reach group also holds the McNemar test of the plan. The registered test of $D$ and the planned tests of the
second human check and of the perfect picker are not in the table; the text gives them.}
\label{tab:supp-holm}
\end{table}

\section{More judges}
\label[appendix]{sec:supp-judge2}

\begin{table*}[t]
\centering\small
\setlength{\tabcolsep}{4pt}
\begin{tabular}{@{}l ccc ccc@{}}
\toprule
 & \multicolumn{3}{c}{First human check} & \multicolumn{3}{c}{Second human check} \\
\cmidrule(lr){2-4}\cmidrule(l){5-7}
Judged by & $D$ (points) & Correct (\%) & Wrong (\%) & $D$ (points) & Correct (\%) & Wrong (\%) \\
\midrule
People & \m5.5 \ci{\m10.3}{\m0.8} & -- & -- & \m6.5 \ci{\m11.1}{\m2.0} & -- & -- \\
Detector judge & \m10.8 \ci{\m15.6}{\m6.0} & 78.9 / 89.7 & 13.6 / 9.4 & \m10.3 \ci{\m15.1}{\m5.5} & 75.6 / 86.7 & 16.5 / 6.7 \\
VLM judge (Qwen3-VL-8B) & \p0.5 \ci{\m3.3}{\p4.3} & 95.0 / 95.0 & 34.7 / 18.8 & \p3.5 \ci{\m0.3}{\p7.5} & 93.3 / 94.2 & 40.9 / 6.7 \\
Second VLM judge (Idefics3-8B) & \m3.8 \ci{\m7.8}{\p0.3} & 86.4 / 92.1 & 32.2 / 17.7 & \m0.8 \ci{\m5.0}{\p3.5} & 85.9 / 90.0 & 35.7 / 10.1 \\
Larger VLM judge (Qwen3-VL-32B) & \m1.5 \ci{\m5.5}{\p2.8} & 90.7 / 94.4 & 33.1 / 14.6 & \m0.8 \ci{\m5.0}{\p3.8} & 89.8 / 94.2 & 35.7 / 7.9 \\
\bottomrule
\end{tabular}
\caption{\textbf{More VLM judges on the images of the two human checks.} $D$: the pick of the 8B VLM picker among
16 base images minus one RL image, in points (negative: RL is better), with 95\% intervals. Correct (Wrong): of the
images that people label correct (wrong), the share that the judge passes, given as best-of-16 / RL. A fair judge
gives two equal numbers in each pair. All three VLM judges pass far more of the wrong best-of-16 images than of the
wrong RL images, so all three report a smaller gap than people. These checks were not planned.}
\label{tab:supp-judge2}
\end{table*}

This section judges the images of both human checks again with three more judges: a VLM judge of another
family, a larger VLM judge of the same family as ours, and our detector judge with a colour classifier that shares
nothing with the reward. None was planned; we added them after the main analysis.

\paragraph{A second VLM judge.}
Our VLM judge hides much of the gap that people see, because it passes many of the wrong images that the VLM picker
selects (\cref{sec:analysis-judges}). The language model of the picker is also a Qwen3 model, so this could be a quirk
of one family of models. We therefore judged the images of both human checks again with a VLM of another family,
Idefics3-8B~\cite{laurencon2024idefics3}. It was trained by another group, and its language model, Llama~3.1, is not a Qwen model. It asks exactly the questions of our VLM judge and passes an image
when every answer passes. It judged the same JPEG images that the labellers saw, and we tuned nothing on its results.

\Cref{tab:supp-judge2} gives the result. The second VLM judge is fooled in the same way as the first. Of the
best-of-16 images that people call wrong, it passes 32.2\% and 35.7\% in the two checks, but only 17.7\% and 10.1\% of
the wrong RL images. This difference, 14.5 and 25.5 points, is clear in both checks (3.2 to 25.6 and 16.0 to 35.1).
So the second judge also reports a smaller gap than people: $-3.8$ and $-0.8$ points, where people see $-5.5$ and
$-6.5$. On the position prompts, where people see RL ahead by 37 and 39 points, it sees 16.5 points in both checks,
and our VLM judge only 6.3 and 10.1. So the leniency towards the picks of a VLM picker is not a property of one
model: two VLM judges of different families share it.

The error of the second VLM judge is smaller than that of our VLM judge, and it is clear only in the second check
(5.8 points, 1.0 to 10.6). One reason is that it also passes fewer of the correct best-of-16 images than of the
correct RL images (86.4\% against 92.1\%, and 85.9\% against 90.0\%), so its two errors partly cancel. This second error
tells us something about the detector judge. It lies on the prompts without colours (7.2 points in both checks, 1.5
to 13.0 and 2.0 to 12.5), where the detector judge is also a little uneven (4.8 and 5.4 points,
\cref{sec:supp-other}). On the prompts with colours, where the detector judge is most uneven, the second VLM judge is
not (3.7 and 0.5 points, both within chance). The smaller unevenness on the other prompts appears in both judges, so
it may come from the picked images themselves.

\paragraph{A larger VLM judge.}
Our VLM judge has 8B parameters, so a larger judge might see the errors that it misses. We therefore also judged the
same images with Qwen3-VL-32B, a judge four times as large and of the same family, with exactly the code, the
questions and the pass rule of our VLM judge. It is lenient in the same way (\cref{tab:supp-judge2}). Of the
best-of-16 images that people call wrong, it passes 33.1\% and 35.7\% in the two checks, but only 14.6\% and 7.9\% of
the wrong RL images, a difference of 18.5 points (7.8 to 28.9) and 27.8 points (18.2 to 37.6). So it also reports a
smaller gap than people: $-1.5$ and $-0.8$ points, where people see $-5.5$ and $-6.5$. On average over both checks it
hides 4.9 points of the gap (1.3 to 8.7). A larger judge of the same kind therefore does not remove the problem.
Unlike the two 8B judges, it is also somewhat uneven on the prompts with colours (8.5 and 8.4 points,
\cref{tab:supp-colour-swap}), though much less than the detector judge.

\begin{table}[t]
\centering\small
\setlength{\tabcolsep}{4pt}
\begin{tabular}{@{}lcccc@{}}
\toprule
 & \multicolumn{2}{c}{Prompts with colours} & \multicolumn{2}{c}{Other prompts} \\
\cmidrule(lr){2-3}\cmidrule(l){4-5}
Judge & First & Second & First & Second \\
\midrule
Detector, CLIP colour & \p22.8 & \p23.1 & \p4.8 & \p6.1 \\
Detector, SigLIP colour & \p20.6 & \p12.6 & \p4.8 & \p6.1 \\
VLM judge & \p0.6 & \p1.9 & \p0.6 & \p1.2 \\
Second VLM judge & \p3.7 & \p0.5 & \p7.2 & \p7.2 \\
Larger VLM judge & \p8.5 & \p8.4 & \p1.2 & \p2.4 \\
\bottomrule
\end{tabular}
\caption{\textbf{Where each judge is uneven.} Of the images that people label correct, the share that a judge passes
for RL minus the share for best-of-16, in points, on the 167 prompts with colours and the 231 others, in the first
and the second human check. A fair judge gives 0. Detector: our detector judge on the JPEG images that people saw,
with the CLIP colour classifier of the reward or with SigLIP; the two differ only on prompts with colours. On the
original images, the verdicts of the paper give 21.1, 21.6, 4.8 and 5.4 (\cref{sec:supp-other}). The
detector judge is uneven on colours with either classifier, the larger VLM judge less, and the two 8B VLM judges
not clearly. These checks were not planned.}
\label{tab:supp-colour-swap}
\end{table}

\paragraph{A detector judge with another colour classifier.}
Our detector judge checks colours with the CLIP classifier of the GenEval reward. To see whether this shared part
causes its bias, we kept everything else of the judge (the OWLv2 boxes, the thresholds and the rules) and replaced
the classifier with SigLIP~\cite{zhai2023siglip}, a model trained elsewhere. With the classifier of the reward, the judge gives the
same verdict as in the paper for all but six images, which differ because people saw JPEG copies.
With SigLIP, the judge is still uneven on the prompts with colours (\cref{tab:supp-colour-swap}). There it passes
67.5\% of the correct best-of-16 images and 88.1\% of the correct RL images in the first check (20.6 points, 11.8 to
29.7), and 69.7\% against 82.3\% in the second (12.6 points, 3.5 to 21.7). With the classifier of the reward, the
same images give 22.8 and 23.1 points. So the weights of the reward explain at most a part of the bias. A colour
check of the same kind, which names the colour of each detected object, favours RL also with a classifier trained
elsewhere. The VLM judges, which see the whole image when they judge a colour, are much less uneven or not clearly
uneven. One reading is that RL
learned colours that such checks read easily; we did not test this further. On average over both checks, the
detector judge with SigLIP still makes the gap larger than people see, by 3.9 points (0.3 to 7.5).

\section{A picker that knows what people would say}
\label[appendix]{sec:supp-search}

\begin{table*}[t]
\centering\small
\setlength{\tabcolsep}{8pt}
\begin{tabular}{@{}lccc@{}}
\toprule
 & All prompts (398) & Prompts without a position (319) & Position prompts (79) \\
\midrule
\multicolumn{4}{@{}l}{\emph{Picker that knows the person's verdict, among $N$ base images}} \\
$N=1$ & 58.0 & 65.8 & 26.6 \\
$N=2$ & 64.8 & 73.4 & 30.4 \\
$N=3$ & 68.3 & 76.2 & 36.7 \\
$N=4$ & 72.6 & 80.6 & 40.5 \\
$N=5$ & 74.4 & 82.4 & 41.8 \\
$N=6$ & 76.1 & 83.4 & 46.8 \\
$N=7$ & 77.9 & 85.3 & 48.1 \\
$N=8$ & 79.1 & 85.9 & 51.9 \\
\midrule
Perfect picker of the paper, $N=8$ & 69.8 & 76.2 & 44.3 \\
One new RL image & 79.1 & 75.9 & 92.4 \\
\midrule
\multicolumn{4}{@{}l}{\emph{This picker minus \dots}} \\
one RL image, $N=4$ & \m6.5 \ci{\m11.1}{\m1.8} & \p4.7 \ci{\p0.6}{\p9.1} & \m51.9 \ci{\m63.3}{\m40.5} \\
one RL image, $N=8$ & 0.0 \ci{\m4.5}{\p4.5} & \p10.0 \ci{\p6.0}{\p14.1} & \m40.5 \ci{\m51.9}{\m29.1} \\
the perfect picker of the paper, $N=8$ & \p9.3 \ci{\p6.5}{\p12.3} & \p9.7 \ci{\p6.6}{\p13.2} & \p7.6 \ci{\p2.5}{\p13.9} \\
\bottomrule
\end{tabular}
\caption{\textbf{A picker that knows what people would say.} Share of prompts (\%) for which at least one of the
first $N$ base images is correct, as judged by a person. No picker can do better by the verdict of that person. The
two middle rows repeat \cref{tab:supp-humano}: the perfect picker of the paper, which uses the two automatic judges,
and one new RL image, both judged by people. The last rows give differences in points with 95\% intervals; these
were not planned tests. With 8 base images, this picker is level with one RL image on all prompts, ahead without a
position and far behind on position prompts.}
\label{tab:supp-search}
\end{table*}

The paper finds that, by our judges, one RL image is worth about 8 base images with a perfect picker (\cref{sec:results-rate}). The
perfect picker of the paper trusts the two automatic judges, and people find its choice among 8 base images 9.3 points
behind one RL image (\cref{sec:supp-humano}). There are two possible reasons. The 8 images may contain no correct image, or
they may contain one that the judges do not find. The human check of the perfect picker had a second part that tells
the two apart. In it, a person looks through the base images of a prompt for a correct one, as in \cref{fig:person}; we
describe this as a picker that knows what people would say. Most of it was labelled after the last date that its plan had
set for the analysis, and in a different way from the first part. This section gives the details and the limits.

\paragraph{What was labelled.}
When a labeller failed the first base image of a prompt, the screen showed the next images of the same prompt, one at
a time, up to the eighth. The images came in the order fixed by the plan, and the series stopped at the first image
that the labeller passed. So we know, for every prompt and every $N$ up to 8, whether the first $N$ base images
contain an image that the person calls correct. This is the best-of-$N$ accuracy of a picker that knows what the
person would say. By this person's verdict, no picker can do better. The labellers failed the first image for 167 of
the 398 prompts and judged 827 further images for them. For 84 of these prompts they found a correct image among the
next seven images, and for 83 they found none among all eight. The two labellers found one in 51\% and 50\% of their
series.

\paragraph{When it was labelled.}
The plan described this part from the start, and it was committed before any task or label of the check existed. On
the day of the first part the labellers reached only 26 of its images. They labelled the other 801 six days later.
This was one day after the last date that the plan had set for the analysis, so we
report this part as a late addition (\cref{sec:supp-rules}). Nothing else changed. The same two labellers used the
same screen and the same written rules, the images and their order were those of the plan, and the labellers had seen
no result of any check. The labels and the results of the first part are as before.

\paragraph{Result.}
\Cref{tab:supp-search} gives the result. With 8 base images, a picker that knows the person's verdict is correct for
79.1\% of the prompts. One RL image is correct for 79.1\% as well. Both numbers are 315 of the 398 prompts, so the
difference is 0.0 points ($-4.5$ to $+4.5$). With 4 base images the picker is still 6.5 points behind ($-11.1$ to
$-1.8$). So, with a picker that knows the person's verdict, the base model reaches one RL image at about 8 images,
which is where our series end. This is close to the exchange rate of the paper, 7.5 by the automatic judges, but it
is less certain. In 48\% of the bootstrap rounds the curve does not reach the RL image within 8 images, so the rate
of this picker may also lie above 8; our series cannot tell by how much.

The perfect picker of the paper does worse with the same 8 images: people find its choice correct for 69.8\% of the
prompts. The 9.3 points between the two pickers (6.5 to 12.3) are 37 prompts on which the 8 images contain a correct
image, but the choice by the judges is wrong. So, by people and over all prompts, the base model draws a correct
image among 8 about as often as one RL image is correct. On average, the perfect picker of the paper falls behind RL
because the judges do not always find that image, and not because the image is missing.

This does not contradict \cref{sec:supp-humano}. There we compared the two graders on the one image that the judges
choose, and their errors nearly cancel (68.8\% by the judges, 69.8\% by people). Here we ask what each grader can
find among the 8 images. The judges find an image to pass for 68.8\% of the prompts, and people for 79.1\%. The
judges also pass the RL image for fewer prompts than people do (66.8\% against 79.1\%). So, over all prompts, the two
models are about level at 8 images when the same side picks and grades: by the judges (68.8\% against 66.8\%) and by
people (79.1\% against 79.1\%). Of these combinations, the base model falls behind only when the judges pick and
people grade, as in the first part.

\paragraph{Planned splits.}
The plan named two splits. As in the first part, they show that the average mixes two different cases. On the 319
prompts without a position, the picker that knows the person's verdict is level with one RL image at 3 base images
(76.2\% against 75.9\%), and at 8 it is 10.0 points ahead (6.0 to 14.1). On the 79 position prompts it reaches only
51.9\% with 8 images, where one RL image is correct for 92.4\% ($-40.5$ points, $-51.9$ to $-29.1$). The perfect
picker of the paper needs 32 images to reach the same 51.9\% (\cref{tab:supp-humano}). For 38 of these 79 prompts
none of the 8 base images is correct, so no picker could have helped. Without a position the curve reaches the RL
image within 8 images in every bootstrap round, and with a position in none. The second split is by prompt set. On the 258
known-object prompts the picker reaches one RL image at 7 base images (70.2\% against 69.0\%), and at 8 the two are
level ($+3.1$ points, $-3.5$ to $+9.7$). On the 140 new-object prompts, where one RL image is correct for 97.9\% of the
prompts, it is still 5.7 points behind at 8 images ($-10.7$ to $-1.4$).

\paragraph{What this adds.}
With the perfect picker of the paper, people confirm the exchange rate of about 8 only for prompts without a position
(\cref{sec:results-rate}), and \cref{sec:supp-humano} finds that, over all prompts, it holds only by our judges. This
section shows what a person can find among the same images, and it makes the two cases sharper. Without a position,
about 3 base images contain a correct one as often as one RL image is correct, so there the limit of best-of-$N$ is
the picker. With a position, 8 base images are far from enough even for the best possible picker, because for about
half of these prompts the base model draws no correct image among 8. So the lead of RL on position prompts does not
come from the picker or from the judges. It is still a matter of how often the base model draws a correct image:
with many more images it solves some of these prompts (\cref{sec:results-reach}). Over all prompts the two cases
average to a rate of about 8, with the uncertainty given above.

\paragraph{Limits.}
This part is weaker evidence than the first part, for four reasons. First, the two models were not labelled in the
same way. The RL image of a prompt was judged once, among images of other prompts. A base image was judged in a
series of the same prompt that stops at the first pass, so the base model had up to eight chances of a pass and the
RL model had one. If a labeller sometimes passes a wrong image, such mistakes add up over a series and favour the
base model. We cannot say how large this effect is. Second, each series was judged by one person, and no second
person saw the images that ended a series. In the first part, the same person had already judged 51 images of these
series, as the first image that both judges pass. Of these 51, 49 received the same label again, in most cases six days
later. But these are
images that the judges pass, so they say little about wrong images that a person might pass. Third, all images of
this part come from the base model. The labellers were not told so, but in a series they saw several images of one
prompt in a row. Fourth, the plan asked for no test of this part, so the intervals in this section only describe the
data; they use the same bootstrap over prompts as the planned test. The series also end at the eighth image, so we
cannot say what such a picker reaches with more images.